\documentclass[10pt,twocolumn,letterpaper]{article}

\PassOptionsToPackage{table}{xcolor}
\usepackage[pagenumbers]{iccv_2025/iccv}
\usepackage{times}
\usepackage{epsfig}
\usepackage{amssymb}

\definecolor{iccvblue}{rgb}{0.21,0.49,0.74}
\usepackage[pagebackref,breaklinks,colorlinks,allcolors=iccvblue]{hyperref}

\usepackage[utf8]{inputenc} 
\usepackage[T1]{fontenc}    
\usepackage{url}            
\usepackage{booktabs}       
\usepackage{amsfonts}       
\usepackage{nicefrac}       
\usepackage{microtype}      
\usepackage{xcolor}         
\usepackage{xspace}
\usepackage{graphicx}
\usepackage{tabularx}
\usepackage{multirow}
\usepackage{float}
\usepackage{makecell}
\usepackage{amsmath}
\usepackage{wrapfig}
\usepackage{bm}
\usepackage{subcaption}
\usepackage{algorithm}
\usepackage{algpseudocode}
\usepackage[capitalize,noabbrev]{cleveref}

\newcommand{\mypara}[1]{\vspace{0pt}\noindent\textbf{#1}}
\definecolor{tealblue}{rgb}{0.21, 0.46, 0.53}
\definecolor{darkseagreen}{rgb}{0.56, 0.74, 0.56}
\definecolor{electricviolet}{rgb}{0.56, 0.0, 1.0}

\begin{document}

\title{NuiScene: Exploring Efficient Generation of Unbounded Outdoor Scenes}

\author{Han-Hung Lee\textsuperscript{1} ~ Qinghong Han\textsuperscript{1} ~ Angel X. Chang\textsuperscript{1,2} \\
\textsuperscript{1}Simon Fraser University, ~ \textsuperscript{2}Canada CIFAR AI Chair, Amii\\
\texttt{\{hla300, qha32, angelx\}@sfu.ca}\\
\href{https://3dlg-hcvc.github.io/NuiScene/}{https://3dlg-hcvc.github.io/NuiScene/}
}

\twocolumn[{
\maketitle
\captionsetup{type=figure}
\centering
\vspace{-3em}
\includegraphics[width=0.95\textwidth,height=4.3cm]{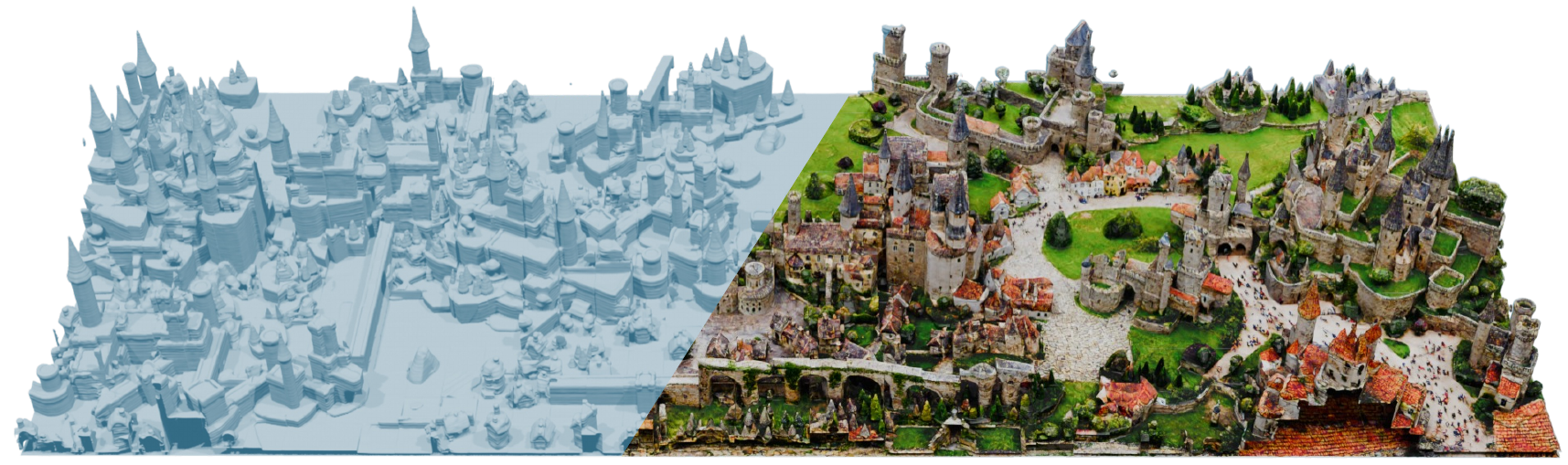}
\captionof{figure}{Our model enables efficient, unbounded generation of large outdoor scene geometry. Scenes are textured with SceneTex~\cite{chen2024scenetex}.}
\label{fig:teaser}
\vspace{1em}

}]

\begin{figure*}
\centering
\vspace{-1em}
\includegraphics[width=\linewidth]{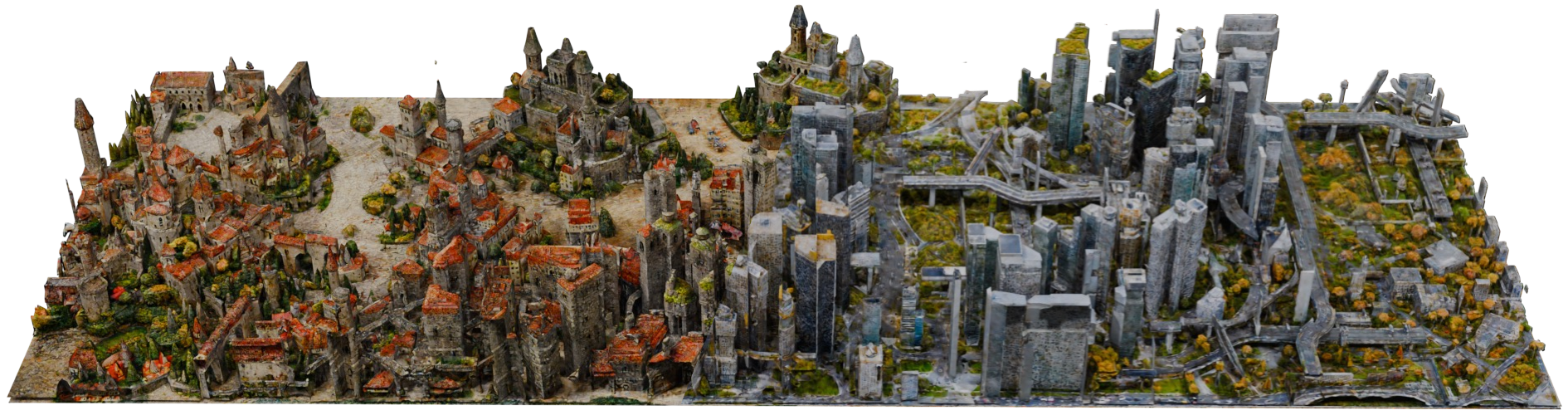}
\caption{A scene generated by our model, trained on multiple scenes. We texture the scene using SceneTex~\cite{chen2024scenetex}. As shown, our model combines elements from different scenes in the dataset such as castles and skyscrapers.}
\label{fig:stitchscene}
\end{figure*}

\begin{abstract}
    In this paper, we explore the task of generating expansive outdoor scenes, ranging from castles to high-rises. Unlike indoor scene generation, which has been a primary focus of prior work, outdoor scene generation presents unique challenges, including wide variations in scene heights and the need for a method capable of rapidly producing large landscapes. To address this, we propose an efficient approach that encodes scene chunks as uniform vector sets, offering better compression and performance than the spatially structured latents used in prior methods. Furthermore, we train an explicit outpainting model for unbounded generation, which improves coherence compared to prior resampling-based inpainting schemes while also speeding up generation by eliminating extra diffusion steps. To facilitate this task, we curate NuiScene43, a small but high-quality set of scenes, preprocessed for joint training. Notably, when trained on scenes of varying styles, our model can blend different environments, such as rural houses and city skyscrapers, within the same scene, highlighting the potential of our curation process to leverage heterogeneous scenes for joint training.
\end{abstract}

\section{Introduction}

Large-scale outdoor scene generation is crucial for many applications such as enabling immersive world-building for open-world video games, cinematic CGI environments, and VR simulations. Traditionally, crafting these vast outdoor environments required extensive manual labor and significant resources. Fortunately, recent advances in AI-driven content generation techniques offer the potential to significantly enhance accessibility and efficiency for this task. 

Following the trend of diffusion models in image generation—such as Stable Diffusion~\cite{rombach2022high}—generative models for creating standalone 3D objects~\cite{xiang2024structured,lee2024text} are also rapidly maturing. The success of such models is partly driven by large-scale datasets like Objaverse~\cite{deitke2023objaverse}. However, large-scale scene generation remains limited, with many works focused on indoor environments~\cite{ju2024diffindscene, wu2024blockfusion, meng2024lt3sd, bokhovkin2024scenefactor}. While some methods utilize semantic car datasets~\cite{lee2023diffusion, liu2024pyramid, lee2024semcity} or CAD models~\cite{lin2023infinicity}, these datasets lack high-quality geometry and have limited scene variation and height. Compared to previous methods, we tackle a difficult task with three key challenges: 1) enabling fast and efficient generation of outdoor scenes with varying heights, 2) generating landscapes with varying styles into a cohesive and continuous scene, 3) curating a dataset that facilitates exploration into such methods and supports joint training of heterogeneous scenes.

\begin{figure}
\centering
\includegraphics[width=\linewidth]{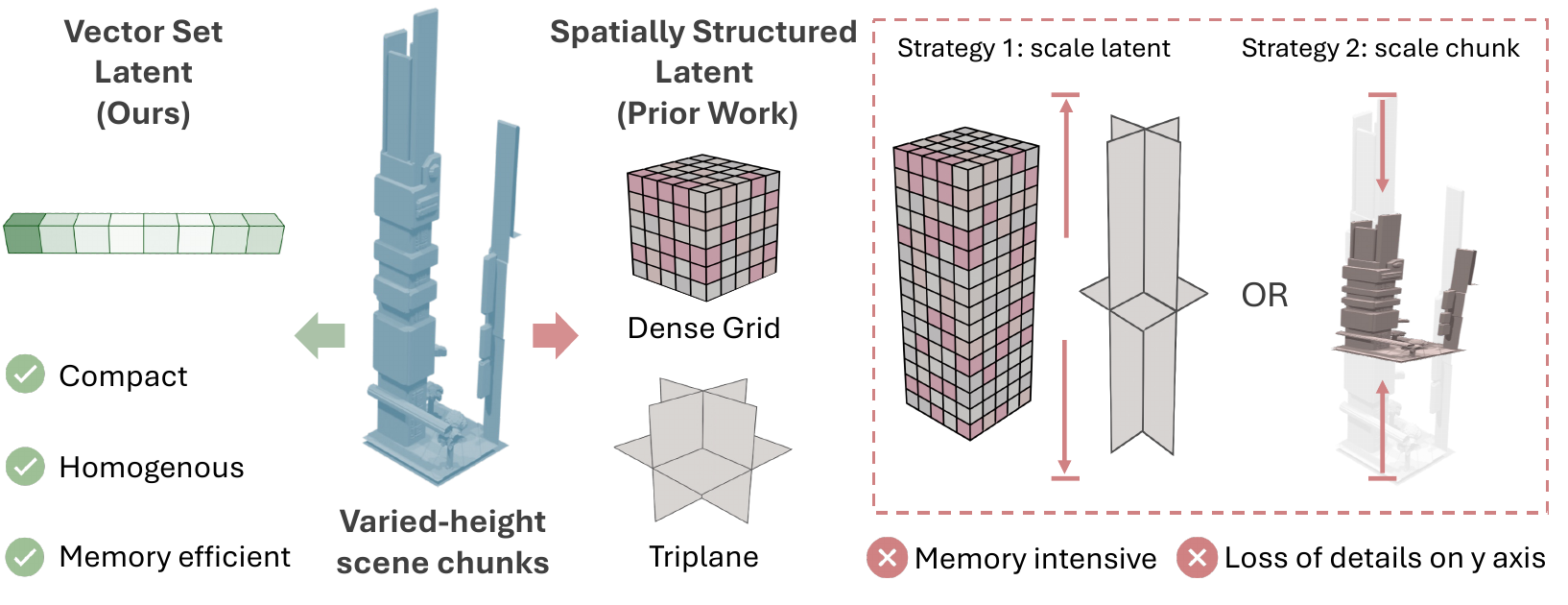}
\caption{
To accommodate scene chunks of varying heights, vector sets offer a compact and uniform representation, whereas prior work relies on spatially structured latents that require scaling either the latent resolution or the chunk itself for compatibility, resulting in high memory usage or loss of detail.
}
\label{fig:latent_compare}
\end{figure}

Previous works on generating unbounded indoor scenes using 3D-Front~\cite{fu20213d} assume scenes can be divided into equal-sized cubes, and learn spatially structured latents like triplanes~\cite{wu2024blockfusion} or dense feature grids~\cite{meng2024lt3sd} for compression. This approach struggles with outdoor scenes featuring buildings like skyscrapers with significantly different heights. As shown in~\cref{fig:latent_compare}, naively scaling these representations can result in intensive memory usage. Rescaling the scene via normalization will also degrade details. To address this, we compress scene chunks into vector sets~\cite{zhang20233dshape2vecset}, achieving better compression than triplanes for more efficient training and higher quality generations. Additionally, we train our diffusion model for explicit outpainting, further improving inference speed over methods~\cite{wu2024blockfusion, meng2024lt3sd} utilizing resampling-based inpainting~\cite{lugmayr2022repaint}.
Furthermore, prior methods focus on uniform datasets, such as indoor or urban driving scenes. In contrast, we aim to explore generative models that can bring together scenes with different context such as castles and cities as demonstrated by our model in~\cref{fig:stitchscene}.

Another key limitation is the lack of openly available high quality outdoor scenes for training. 3D-Front~\cite{fu20213d} is used by many prior works to train unbounded indoor scene generation. However, it lacks height diversity and non-furniture geometry. Semantic driving datasets~\cite{wilson2022motionsc, behley2019semantickitti} and urban reconstructions~\cite{riemenschneider2014learning, gao2021sum, gao2024cus3d} offer limited mesh quality. Objaverse~\cite{deitke2023objaverse} includes many scenes that could facilitate training for this task. However, several challenges remain. Notably, these meshes lack a unified global scale, and feature highly varied ground geometries—ranging from very thick to thin planes—making it difficult to train a generative model effectively as a homogeneous dataset. To address this, we curate a small set of scenes from Objaverse, clean ground geometries and establish a unified scale, enabling the development of methods that handle (1) diverse scene height distributions and (2) outdoor geometry from varied sources and styles. We show that our dataset pipeline effectively enables joint training of heterogeneous scenes in a unified model.

In summary, NuiScene provides the following contributions: \textbf{Efficient Representation.} We propose using vector sets to encode chunks of varying sizes into a uniform representation, enhancing both training efficiency and scene generation quality. \textbf{Outpainting Model.} Unlike previous methods relying on slow resampling-based outpainting, we train our diffusion model to explicitly outpaint by conditioning on previously generated whole chunks, leading to faster inference time generation. \textbf{Cross Scene Generation.} We demonstrate that our efficient NuiScene model, trained on four curated scenes, can blend scenes of different styles (e.g., medieval, city), highlighting both the model's capability and effectiveness of our curation process. \textbf{NuiScene43.} A dataset of 43 moderate to large sized scenes from Objaverse, with cleaned ground geometries and unified scales. This enables joint training on heterogeneous scenes, supporting the development of methods for unbounded scene generation with 1) diverse scene heights, and 2) various styles.  

\section{Related Work}

\begin{figure*}
\centering
\includegraphics[width=0.95\linewidth]{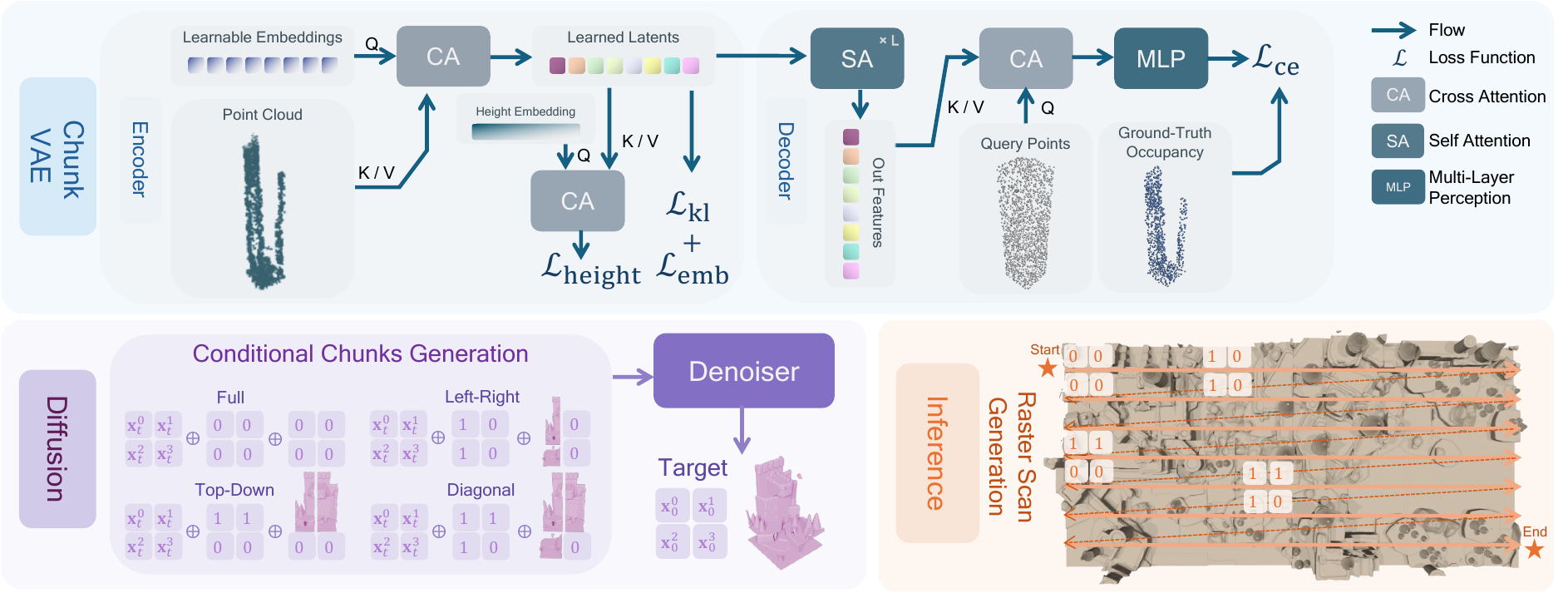}
\caption{Overview of our VAE and diffusion models. For the VAE we use vecset~\cite{zhang20233dshape2vecset} for latent compression. For the diffusion model we train a conditional outpainting model with four different settings to allow for fast generation in a raster scan pattern during inference.}
\label{fig:model}
\end{figure*}

\mypara{3D Scene Gen with T2I Priors.} Recently, methods leveraging priors from text-to-image (T2I) diffusion models~\cite{rombach2022high} and monocular depth estimation methods~\cite{bhat2023zoedepth, lasinger2019towards, ke2024repurposing} have gained popularity for generating 3D scenes from text or images~\cite{chung2023luciddreamer, shriram2024realmdreamer, yu2024wonderworld, fridman2023scenescape}. These approaches use T2I models to generate RGB images and depth prediction models to extract geometric information, which is then used to distill or optimize 3D geometry. This enables continuous outpainting and unbounded scene generation by shifting the camera and synthesizing new regions. While the strong priors of T2I models allow for open-domain scene generation, accumulated errors in depth predictions lead to geometric distortions and limitations in long-range consistency.

\mypara{3D Bounded Single Exemplar Scene Gen.} Several works focus on generating 3D scenes from a single exemplar~\cite{wang2024singrav, karnewar20223ingan, wu2022learning, li2023patch}, follow a coarse-to-fine progressive approach by capturing patch statistics across scales. Similar to SinGAN~\cite{shaham2019singan}, SinGRAV~\cite{wang2024singrav}, 3INGAN~\cite{karnewar20223ingan}, and \citet{wu2022learning} adopt a pyramid of GANs, where discriminators operate on smaller patch sizes to enforce local consistency while allowing diverse global compositions. 
\citet{li2023patch} introduces EM-like optimization for patch matching and blending across different scales. Although these methods generate scenes with varying compositions and aspect ratios, they are limited to bounded generation, and producing an entire scene at once is memory intensive for larger scenes.

\mypara{3D Unbounded Scene Gen.} 
One common framework for 3D unbounded scene generation starts by dividing scenes into smaller chunks and using an autoencoder to compress them into latent representations. A diffusion model is then trained on the scene chunk latents. By overlapping the current chunk with previously generated ones, these methods can enable continuous outpainting by using resampling-based inpainting methods like RePaint~\cite{lugmayr2022repaint}.

Unbounded outdoor scene generation has been explored using synthetic and real-world semantic driving datasets like CarlaSC~\cite{wilson2022motionsc} and SemanticKITTI~\cite{behley2019semantickitti}. 
While these datasets enable relatively large-scale scene generation, they lack high-quality geometry.
SemCity~\cite{lee2024semcity} follows the approach outline above and uses triplanes as the latent representation. PDD~\cite{liu2024pyramid} employs a pyramid discrete diffusion model on raw semantic grids directly with a scene subdivision module that conditions generation on overlapping regions. Unlike PDD which conditions on overlapping regions using dense grids, our approach conditions on entire neighboring chunks with richer context. These methods suffer from limited geometric quality, and height variation due to the dataset.

For unbounded indoor scene generation, methods typically train on 3D-Front~\cite{fu20213d} and assume cubic scene chunks. BlockFusion~\cite{wu2024blockfusion}, like SemCity, compresses scene chunks into triplanes. It then uses another VAE to further compress the triplanes into a lower resolution, reducing redundancy in the representation to improve diffusion learning. LT3SD~\cite{meng2024lt3sd} encodes TUDF chunks into multi-level dense feature grid latent trees, organizing both its autoencoder and diffusion models in a hierarchical manner for coarse-to-fine chunk generation. While these representations perform well for indoor scenes with fixed ceiling heights, they struggle to scale to outdoor environments with significant height variations, as shown in~\cref{fig:latent_compare}. To tackle this, we use a vector set representation which compresses chunks into a uniform size, offering better compression and performance, and enable faster generation with a outpainting model.

Other methods take a different approach to unbounded outdoor generation. CityDreamer~\cite{xie2024citydreamer} and InfiniCity~\cite{lin2023infinicity} leverage top-down views for unbounded generation but are limited to cityscapes with low-resolution geometry. SceneDreamer~\cite{chen2023scenedreamer} and Persistent Nature~\cite{chai2023persistent} are promising directions to generate unbounded 3D worlds from only 2D images. However, without explicit 3D supervision, the resulting geometry lacks detail.

\section{Method}

We represent a large outdoor scene as a collection of local scene chunks, where a VAE learns to compress individual chunks into vector sets, and a diffusion model generates neighboring chunks in a $2\times2$ grid using the VAE's learned representation. By combining both models, we enable continuous and unbounded outdoor scene synthesis. Our training framework (see \cref{fig:model}) follows the Latent Diffusion Model (LDM)\cite{rombach2022high}.

We outline the dataset curation process and the sampling of training chunks in~\cref{sec:method-dataset}. To explore efficient scene generation, we experiment with triplane and vector set latents. The VAE backbone and the process of obtaining occupancies from these representations are described in~\cref{sec:model-vae}. For fast inference, we train an explicit outpainting diffusion model, detailed in~\cref{sec:model-diffusion}. Optionally, textures can be generated using SceneTex~\cite{chen2024scenetex} as shown in~\cref{sec:model-texturing}.

\subsection{Dataset}
\label{sec:method-dataset}

\mypara{Filtering and Preprocessing.} We begin by filtering Objaverse~\cite{deitke2023objaverse} using multi-view embeddings from DuoduoCLIP~\cite{lee2024duoduo} to select 43 high-quality scenes. Next, we establish a consistent scale by labeling scenes with relative scales. We then convert each scene into an occupancy grid with a resolution multiplied by its labeled relative scale and generate meshes using the marching cubes algorithm for point cloud sampling. To ensure uniform ground geometry across scenes, we ensure a uniform thickness below the ground level for each scene. For more details, see the supplement.

\begin{figure}
\centering
\vspace{-1em}
\includegraphics[width=\linewidth]{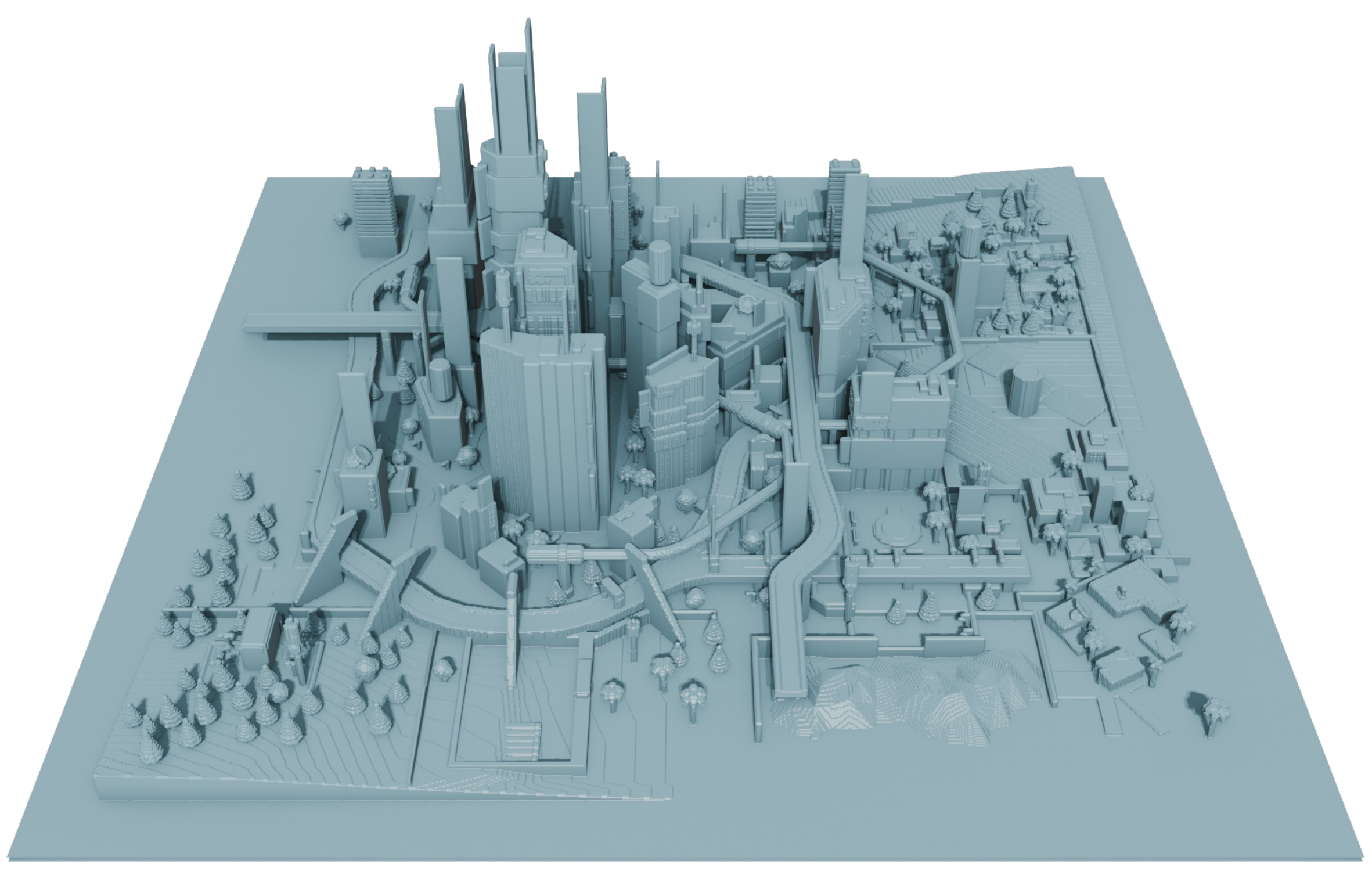}
\caption{A sample scene processed using our curation pipeline. The occupancy is processed with marching cubes to get the mesh.}
\label{fig:gt_single_scene}
\end{figure}

\mypara{Chunk Sampling.} A sample scene processed through our curation pipeline is shown in~\cref{fig:gt_single_scene}. The resulting occupancy grid \verb|occ| has dimensions of ${1059 \times 478\times 1059}$. To facilitate training, we divide the scene into smaller chunks of size $(50, h_\text{vox}, 50)$ along the $x, y, z$ axis, where $y$ represents the height axis. Chunks are sampled via \verb|occ[i-25:i+25,:,k-25:k+25]|, where $i$ and $k$ are sample coordinates along the $x$ and $z$ axis of the scene, respectively. The height of the sampled chunk $h_\text{vox}\leq 478$ is determined by the tallest occupied voxel within each chunk. 

For training, we sample coordinates within the chunk's occupancy grid as $\bm{r}_\text{vox} = (x_\text{vox}, y_\text{vox}, z_\text{vox})$, where $0 \leq x_\text{vox}, z_\text{vox} \le 50$ and $0 \leq y_\text{vox} \le h_\text{vox}$ along with the corresponding occupancy $\tilde{o}$ at each coordinate. The sampled coordinates are then normalized as: $\bm{r} = 2 * (\bm{r}_\text{vox} / \bm{d}) - 1$, where $\bm{d} = (d_x, d_y, d_z)$ is the normalization scale for each axis. The ground truth height of the chunk is computed as $\tilde{h} = 2 * (h_\text{vox} / 50) - 1$. The chunk’s point cloud is also normalized using the same procedure with $\bm{d} = (50, 50, 50)$.

\begin{figure*}
\centering
\vspace{-1em}
\includegraphics[width=\linewidth]{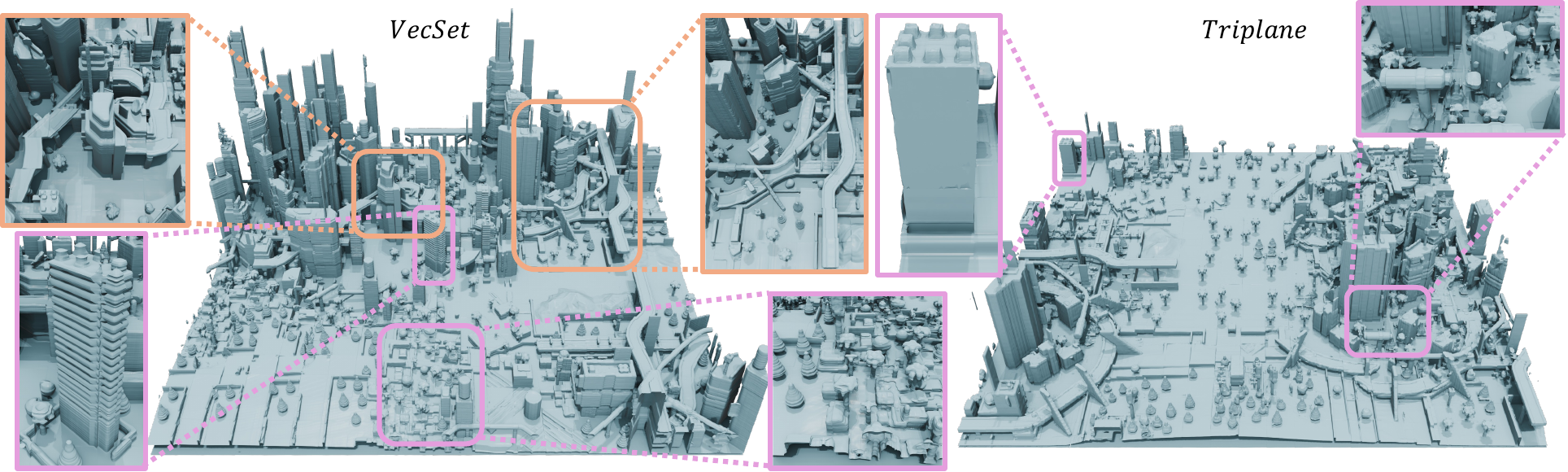}
\caption{We show generation results for the vector set and triplane baseline trained on a single scene. {\color{orange}Orange} boxes and zoomed-in renders highlight chunk outpainting coherence, while {\color{violet}purple} boxes illustrate scene geometry (zoom in for details). Zoomed-in renders may also have adjusted camera angles for clarity. The triplane model struggles with building details and introduces noisy artifacts, whereas the vector set model has better fidelity. While models can generate continuous chunks with smoothly connected roads and pipes (orange box, right), they sometimes fail to maintain coherence across chunks, leading to broken bridges (orange box, left).}
\label{fig:init-results}
\end{figure*}

\subsection{Chunk VAE}
\label{sec:model-vae}

\subsubsection{Encoder}

Our encoder $\mathcal{E}$ compresses sampled scene chunks $P$ consisting of varying heights into a uniform set of vectors. For each chunk $P_i$, we uniformly sample a fixed size point cloud $\bm{p}\in \mathbb{R}^{N_p\times3}$ of $N_p$ points. Following 3DShape2Vecset~\cite{zhang20233dshape2vecset}, we use a cross-attention (CA) layer to aggregate the point clouds into a compact representation, leveraging a learnable set of fixed features with a reduced number of tokens. Next, a fully connected (FC) layer predicts the mean and variance, from which we apply the reparameterization trick~\cite{kingma2013auto} to generate embeddings for the chunks. This results in $\bm{z}^{\bm{p}}=\mathcal{E}(\bm{p})\in \mathbb{R}^{V\times c}$, where $V$ is the number of vectors and $c$ is the channel size.

However, we found that this is prone to posterior collapse~\cite{van2017neural}, a common issue in VAEs. To address this, we introduce an alternative regularization strategy. Specifically, we sample another point cloud $\bm{q}\in \mathbb{R}^{N_p\times3}$ from the same chunk $P_i$ and enforce consistency between their embeddings: $ \mathcal{L}_\text{emb} = (\bm{z}^{\bm{p}} - \bm{z}^{\bm{q}})^2$. This prevents VAE collapse as it prevents the encoder from ignoring its input and enforces similar embeddings for point clouds sampled from the same chunk, with minimal overhead—one extra CA and FC pass.

Furthermore, we predict the chunk’s height from its embedding to prune unnecessary occupancy predictions at inference time. At inference time, the embedding is generated by the diffusion model and thus we do not have access to the ground truth height of each chunk. To account for this, we learn a height embedding $\bm{e}_h\in \mathbb{R}^{1\times c}$ and use it to query the latent representation, yielding the predicted height $\hat{h} = \text{FC}(\text{CA}(\bm{z^{p}}, \bm{e}_h))$. The height loss is calculated as $\mathcal{L}_\text{height}=(\hat{h} - \tilde{h})^2$ where $\tilde{h}$ is the ground truth height.

\subsubsection{Decoder}

Next, we pass the embedding through $L$ self-attention (SA) layers in the decoder to obtain the output features $\bm{f}_\text{out} = \text{SA}^{(L)} \circ \cdots \circ \text{SA}^{(1)}(\bm{z^{p}})$, which are then passed to either the vector set or triplane head to decode the occupancy.

\mypara{Triplane.}
For the triplane head, $\bm{f}_\text{out}$ is reshaped into a triplane and upsampled via deconvolutions, following LRM~\cite{hong2023lrm}, yielding a feature grid $T \in \mathbb{R}^{3 \times h_{\text{tri}} \times w_{\text{tri}} \times c_{\text{tri}}}$. Query points $\bm{r_{tri}}$ as outlined in~\cref{sec:method-dataset} are normalized with $\bm{d}=(50, 50*S, 50)$, where the scale factor $S$ further compresses points along the y-axis. To ensure all coordinates remain within $[-1, 1]$ for valid triplane sampling, we apply clamping. Note that we define $-1$ as the left or bottom bound of the triplane and $1$ as the right or top bound.  Features are extracted via bilinear interpolation from the triplanes, concatenated, and processed by a FC layer to predict occupancies $\hat{o}_r$.

\mypara{Vector Set.} The query point $\bm{r_{vec}}$ is calculated with a normalizing scale of: $\bm{d}=(50, 50, 50)$. Occupancy is predicted via cross-attention: $ \hat{o}_r = \text{FC}(\text{CA}(\bm{f}_\text{out}, \text{PE}(\bm{r_{vec}})))$, where PE is the fourier embedding of the query points.

\mypara{Loss.} We supervise the occupancy with binary cross entropy $\mathcal{L}_\text{ce}=\text{BCE}(\hat{o}_r, \tilde{o}_r)$ with the ground truth occupancy $\tilde{o}_r$. The embedding is supervised with a KL loss $\mathcal{L}_\text{kl}$. See~\citet{zhang20233dshape2vecset} for more details. The total loss is $\mathcal{L}=\lambda_\text{kl} \mathcal{L}_\text{kl} + \lambda_\text{emb} \mathcal{L}_\text{emb} + \lambda_\text{ce} \mathcal{L}_\text{ce} + \lambda_\text{height} \mathcal{L}_\text{height}$, and $\lambda$s are the loss weights. We note that, unlike the triplane method, the vector set allows query points to have ranges beyond $[-1, 1]$. This is because cross-attention and positional encoding can directly handle arbitrary values without requiring them to be within a specific range for valid sampling.

\mypara{Inference Time Decoding.} During inference, since we have no access to the ground truth height $\tilde{h}$, we use the predicted height $\hat{h}$ to prune unnecessary calculations. First, our model predicts $\hat{h}$. Then, we generate all query coordinates $\bm{r}_\text{vox}$ within a volume of shape $(50, 50 * (\hat{h} + 1) / 2, 50)$, where each query point $\bm{r}$ is normalized via $2 * (\bm{r}_\text{vox} / \bm{d}) - 1$. These normalized coordinates query the latent representation for occupancy predictions, and we apply marching cubes to the occupancy volume to reconstruct the chunk’s mesh.

\subsection{Diffusion}
\label{sec:model-diffusion}

Unlike existing models~\cite{lee2024semcity, wu2024blockfusion, meng2024lt3sd} that use RePaint~\cite{lugmayr2022repaint} for outpainting, requiring extra diffusion steps, we propose a more efficient approach. Our outpainting diffusion model generates four chunks at once in a $2\times 2$ grid, conditioning on previously generated chunks using different patterns to handle all scenarios as shown in~\cref{fig:model} for continuous generation in a raster scan order.

We adopt DDPM~\cite{ho2020denoising} for diffusion probabilistic modeling. During training, we sample a quadrant of neighboring chunk latents $\{\bm{z}^0,\bm{z}^1,\bm{z}^2,\bm{z}^3\}$ arranged in a $2\times2$ grid sampled from the scene and encoded by the VAE. The order of chunks is shown in~\cref{fig:model}. At each training step we sample a Gaussian noise $\bm{\epsilon}\sim\mathcal{N}(0,\mathbf{I})$ at a randomly sampled timestep $t\in[1,T]$ which is added to the quadrant of embeddings to obtain noisy latents $\bm{X}_t = \{\bm{x}^0_t, \bm{x}^1_t, \bm{x}^2_t, \bm{x}^3_t\}\in \mathbb{R}^{4\times V\times c}$, where $\bm{x}^i_t \in \mathbb{R}^{V\times c}$ is the noised $i$-th latent at timestep $t$.

Our diffusion model takes noisy latents $\bm{X}_t$ as input, along with a binary mask indicating existing chunks and corresponding conditional embeddings providing context for outpainting. We consider four possible conditioning configurations, each with a corresponding mask $M$ and condition embeddings $Z_\text{cond}$, as illustrated in~\cref{fig:model}. The mask is defined as $M = \{\bm{m}^0, \bm{m}^1, \bm{m}^2, \bm{m}^3\}\in \mathbb{R}^{4\times V\times 1}$, where $M$ has four possible values: $\{\mathbf{0}, \mathbf{0}, \mathbf{0}, \mathbf{0}\}$, $\{\mathbf{1}, \mathbf{0}, \mathbf{1}, \mathbf{0}\}$, $\{\mathbf{1}, \mathbf{1}, \mathbf{0}, \mathbf{0}\}$ or $\{\mathbf{1}, \mathbf{1}, \mathbf{1}, \mathbf{0}\}$, with $\mathbf{0}, \mathbf{1}\in \mathbb{R}^{V\times1}$.
The condition embeddings are constructed based on the corresponding masks $\{\mathbf{0}, \mathbf{0}, \mathbf{0}, \mathbf{0}\}$, $\{\bm{z}^0, \mathbf{0}, \bm{z}^2, \mathbf{0}\}$, $\{\bm{z}^0, \bm{z}^1, \mathbf{0}, \mathbf{0}\}$ and $\{\bm{z}^0, \bm{z}^1, \bm{z}^2, \mathbf{0}\}$ with $\mathbf{0}\in \mathbb{R}^{V\times c}$ to form $Z_\text{cond}$. The mask, conditional embeddings and positional embeddings $\text{PE}$ are concatenated to obtain the condition of the diffusion model:
\begin{equation} 
\bm{C} = M \oplus Z_{\text{cond}} \oplus \text{PE}
\end{equation}

We use a UNet-style Transformer from Craftsman~\cite{li2024craftsman} as the backbone for denoising network $\epsilon_\theta$. The model is then trained to denoise latents via:
\begin{equation}
    \mathbb{E}_{\bm{X}, \bm{C}, \bm{\epsilon}\sim\mathcal{N}(0, \mathbf{I}), t}\left[\|\bm{\epsilon} - \epsilon_\theta((\bm{X}_t\oplus \bm{C}), t)\|^2_2\right]
\end{equation}

During diffusion model training, one of the four configurations is randomly selected, along with the corresponding mask and conditional embeddings. Please see the supplemental for more details on the raster scan generation.

\subsection{Scene Texturing}
\label{sec:model-texturing}

We optionally texture generated scenes using SceneTex~\cite{chen2024scenetex}, modifying camera settings for better coverage of our scenes compared to the default {Spherical Camera}\footnote{\url{https://github.com/daveredrum/SceneTex?tab=readme-ov-file\#cameras}}. Using Blender, we manually keyframe the scene with a fixed camera trajectory and speed, yielding 1.2k to 1.8k keyframes depending on scene shape and scale to use for SceneTex training. Scenes are normalized to $[-1,1]^3$, and we set the camera's field of view to 40\textdegree~with a near clipping distance of 0.01m. Please refer to the supplement for more details.
\section{Experiment}

\subsection{Dataset}

For single-scene experiments, we use the processed scene from~\cref{fig:gt_single_scene} for training. Instead of sampling individual chunks, we sample larger quad-chunks of size $(100, h_{vox}, 100)$, which consists of a 2x2 grid of smaller chunks $(50, h_{vox}, 50)$. Sampling is done along the $x$ and $z$ axes using \verb|occ[i-50:i+50,:,k-50:k+50]|. We sample 100k quad-chunks in total, with 95k used for training and 5k for validation. While these quad-chunks may overlap, they are not identical, as they come from different $x$ and $z$ coordinates. We define the batch size as the number of quad-chunks used for training, with the quad-chunks divided into 4 smaller chunks during VAE training (380k train/20k val). For the diffusion model, the quad-chunks’ point clouds are encoded using the VAE for diffusion training.

For multi-scene experiments, we add three additional scenes in additional to the single scene for training. In total we sample 300k quad-chunks from these scenes.

\subsection{Evaluation Metrics}

\mypara{Reconstruction.} For reconstruction metrics, we report Chamfer Distance (CD), F-Score, and IoU following~\citet{zhang20233dshape2vecset} by comparing ground truth and reconstructed chunks on the validation set. CD and F-Score are computed from 50k sampled points per chunk using a threshold of $2/50$ (the voxel length of sampled chunks). For IoU, we sample 20k occupancy values, with 10k split between occupied and unoccupied regions, along with 10k near-surface points.

\mypara{Generation.} We evaluate the generation quality of our diffusion models using Fréchet PointNet++~\cite{qi2017pointnet++} Distance (FPD) and Kernel PointNet++ Distance (KPD), following~\citet{zhang20233dshape2vecset}. We use the Pointnet++ model from Point-E~\cite{nichol2022point} following~\citet{yan2024object}. Specifically, we sample 10k quad-chunks from the diffusion model, compute their surface point clouds (2048 points), and compare them to 10k sampled quad-chunks from the original scene.

\subsection{Model and Baselines}

We use triplane representations as a baseline for this task, given their widespread use in prior work \cite{lee2024semcity, wu2024blockfusion}. Note, we did not directly adopt BlockFusion’s triplane architecture, as it requires fitting triplanes for all dataset chunks, a process that took thousands of GPU hours on 3D-Front. Instead, we adopt an LRM~\cite{hong2023lrm} style VAE backbone (\cref{sec:model-vae}).

We exclude dense grid methods~\cite{liu2024pyramid, meng2024lt3sd} due to their high memory demands, which scale cubically with resolution, compared to the quadratic growth of triplanes. In our task, the height can be up to about 8.5 times greater than the $x$ and $z$ dimensions for scene chunks in~\cref{fig:gt_single_scene}, further exacerbating memory usage. Additionally, LT3SD~\cite{meng2024lt3sd} requires multi-scale training for both the autoencoder and diffusion models, increasing training time.

\mypara{VAE.} For triplane baselines, we explore configurations with $V=3\times4^2$, varying the normalization scale $S$, and the number of deconvolution layers, which affects output resolution (see~\cref{tab:vae_resource}). For the vector set model, compressing to $V=16$ preserves geometric fidelity while enabling efficient diffusion training. Both representations use a channel size of $c=64$, and a sampled point cloud size of $N_p=4096$. We set the output triplane channel $c_{tri}=40$. The decoder has $L=24$ self-attention layers, and all models are trained for $160$ epochs. Each training iteration samples $4096$ query points per chunk for occupancy supervision.

\mypara{Diffusion.} For all models, we use a 25-layer UNet-like transformer~\cite{li2024craftsman} for diffusion. The triplane diffusion model operates on embeddings from the VAE with the output triplane resolution of $h_{tri} = w_{tri} = 64$ and $S=6$. We train all models for $320$ epochs and a batch size of $192$. The transformer processes tokens of length $4\times V$ as the diffusion model generates a 2x2 grid of chunks at once.

\subsection{Single Scene}
\label{sec:exp_single_scene}

In single-scene experiments, we demonstrate the effectiveness of representing scene chunks as vector sets compared to spatially structured latents like triplanes. VAE reconstruction results (\cref{sec:vae_recon}) and diffusion model comparisons (\cref{sec:diffusion}) show that the vector set representation achieves better compression, leading to improved training efficiency and superior performance over triplanes.

\subsubsection{VAE Reconstruction}
\label{sec:vae_recon}

\cref{tab:vae_recon} presents the quantitative reconstruction results, showing that the vector set model outperforms all triplane settings. We also report whether occupancy queries are pruned using the predicted height $\hat{h}$ or the ground truth height $\tilde{h}$ when computing meshes for CD and F-Score. The similar values in both cases indicate the accuracy of our height prediction.

The triplane models are constrained by the VAE’s output resolution, with performance improving as resolution increases from $3\times32^2$ to $3\times64^2$. However, for tall scenes, coordinate clamping during triplane sampling prevents reconstruction of tall buildings exceeding the scaling factor of $S=6$. Increasing $S$ to $8.5$ (matching the tallest chunk) severely degrades performance due to precision loss from excessive compression along the y-axis. While further increasing resolution to $128^2$ via deconvolution is possible, it requires training on 4 L40S GPUs (\cref{tab:vae_resource}). In contrast, the vector set model provides better compression ($V=16$ vs $V=3\times4^2$) while requiring similar training time and memory as the lowest-resolution triplane model, demonstrating its efficiency in representing scene chunks and superior performance. See additional results in the supplement.

\begin{table}
\centering
\begin{minipage}{0.45\textwidth}
\caption{Comparison of VAE training resources for vector set vs triplane backbones. Training for most experiments was run on 2 L40S GPUs, total batch size and memory across all gpus are reported. The \# Latents is the size of the latent for the VAE backbone and Output Res indicates the triplane size after deconvolution. Increasing the output triplane resolution to $3\times128^2$ requires 4 GPUs for training.}
\vspace{-1em}
\resizebox{\linewidth}{!}
{
\begin{tabular}{@{}lrrrrrr@{}}
\toprule
Method & BS & \# Latents & Output Res & S & Time (hr) & Mem. (GB)\\
\midrule
\multirow{3}{*}{triplane} & 40 & $3\times4^2$ & $3\times32^2$ & 6 & 35.9 & 35.7 \\
& 40 & $3\times4^2$ & $3\times64^2$ & 6 & 58.5 & 55.7 \\
& 40 & $3\times4^2$ & $3\times128^2$ & 6 & 86 & 163.2 \\
\midrule
vecset & 40 & 16 & - & - & 36.1 & 36.6 \\
\bottomrule
\end{tabular}
}
\label{tab:vae_resource}
\end{minipage}
\hspace{1cm}
\begin{minipage}{0.45\textwidth}
\caption{Comparison of difffusion training resources of vector set representation against triplanes. Training across all experiments was run on 1 A6000 GPU. The \# Tokens are the token length for the transformer.}
\vspace{-1em}
\resizebox{\linewidth}{!}
{
\begin{tabular}{@{}lrrrr@{}}
\toprule
Method & BS & \# Tokens & Time (hr) & Mem. (GB)\\
\midrule
triplane & 192 & $4\times3\times4^2$ & 27.6 & 24.4 \\
\midrule
vecset & 192 & $4\times16$ & 11.1 & 10.4 \\
\bottomrule
\end{tabular}
}
\label{tab:diff_resource}
\end{minipage}
\end{table}

\begin{table}
\centering
\caption{Quantitative comparison of reconstruction across different VAE backbones. Here $\hat{h}$ indicates that the predicted height was used for the occupancy prediction and $\tilde{h}$ the ground truth height.}
\vspace{-1em}
\resizebox{\linewidth}{!}
{
\begin{tabular}{@{}llrrrrr@{}}
\toprule
Method & Output Res/S & IOU$\uparrow$ & CD $(\hat{h})\downarrow$ & F-Score $(\hat{h})\uparrow$ & CD $(\tilde{h})\downarrow$ & F-Score $(\tilde{h})\uparrow$\\
\midrule
\multirow{3}{*}{triplane} & $3\times32^2$/$6$ & 0.734 & 0.168 & 0.508 & 0.168 & 0.508 \\
& $3\times64^2$/$8.5$ & 0.805 & 0.105 & 0.705 & 0.105 & 0.705 \\
& $3\times64^2$/$6$ & 0.940 & 0.064 & 0.831 & 0.064 & 0.831 \\
\midrule
vecset & - & $\mathbf{0.989}$ & $\mathbf{0.055}$ & $\mathbf{0.864}$ & $\mathbf{0.055}$ & $\mathbf{0.863}$ \\
\bottomrule
\end{tabular}
}
\label{tab:vae_recon}
\end{table}

\subsubsection{Diffusion}
\label{sec:diffusion}

\begin{figure}
\centering
\includegraphics[width=0.95\linewidth]{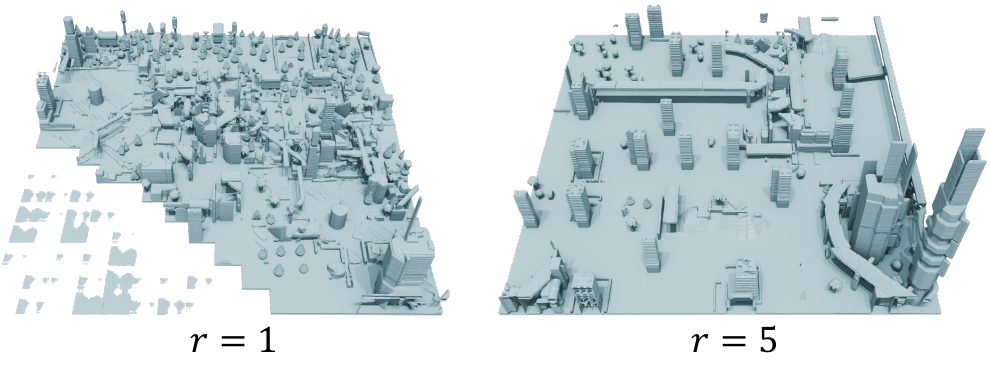}
\caption{We compare generation results using RePaint~\cite{lugmayr2022repaint} for outpainting with resampling steps $r=1$ and $r=5$.}
\label{fig:repaint}
\end{figure}

\begin{table}
\centering
\caption{Comparison of triplane and vecset diffusion models for generated quad-chunks. KPD scores are multiplied $10^3$.}
\vspace{-1em}
\resizebox{0.45\linewidth}{!}
{
\begin{tabular}{@{}lrr@{}}
\toprule
Method & FPD$\downarrow$ & KPD$\downarrow$ \\
\midrule
triplane & 1.406 & 2.589 \\
vecset & $\mathbf{0.571}$ & $\mathbf{0.951}$ \\
\bottomrule
\end{tabular}
}
\label{tab:kpd_fpd}
\end{table}

\begin{figure*}
\centering
\vspace{-6pt}
\includegraphics[width=0.95\linewidth]{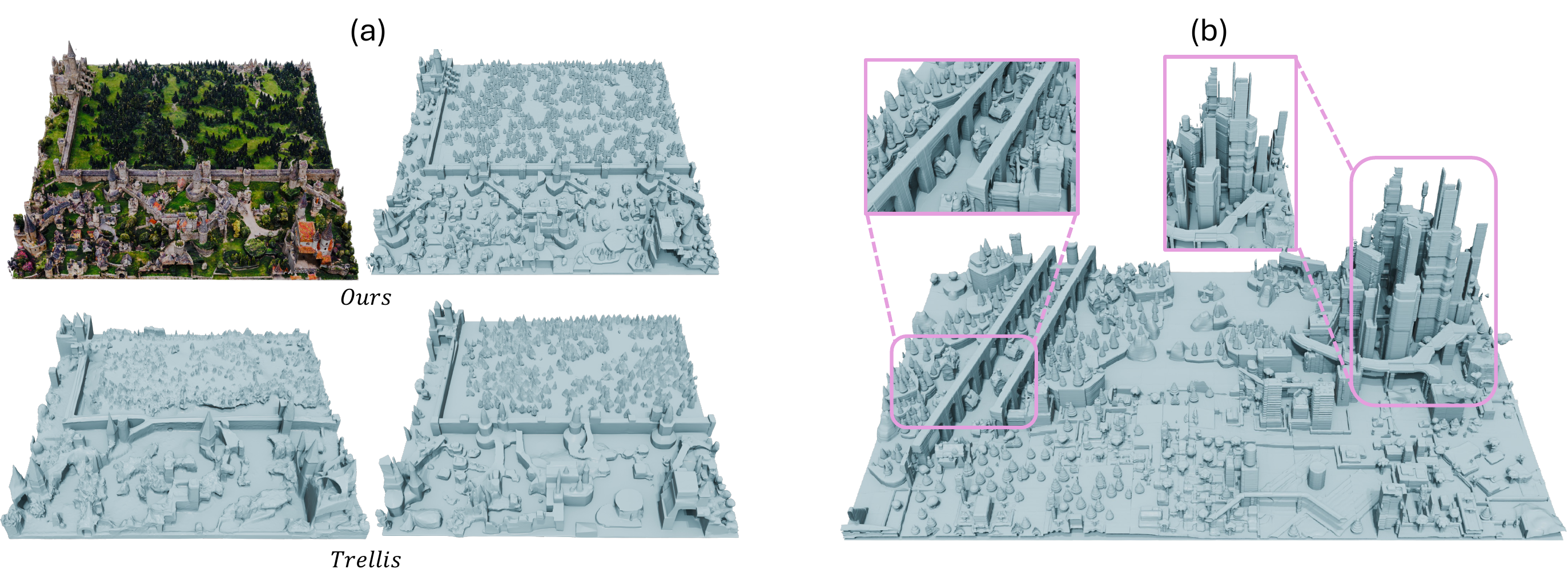}
\caption{We present results from our NuiScene model trained on multiple scenes. In Figure (a), the top-right shows the generated geometry, while the top-left displays the textured model using SceneTex~\cite{chen2024scenetex}. We render these models and feed them into Trellis~\cite{xiang2024structured} for reconstruction on the bottom (zoom in to see details). Our method produces better geometric detail. Figure (b) presents a large scene generated by our model, showcasing its ability to blend elements from different scenes in the dataset.}
\vspace{-1em}
\label{fig:multiscene}
\end{figure*}

\mypara{Quality.} The FPD and KPD scores (\cref{tab:kpd_fpd}) show that the vector set diffusion model outperforms the baseline triplane diffusion model as well. Qualitative results for a 21x21 chunk scene (\cref{fig:init-results}) demonstrate the vector set model’s ability to generate finer details, especially for buildings, whereas the triplane model struggles with details and introduces noisy artifacts which contributes to the higher FPD and KPD scores, likely due to compression of query points along the y axis. Additionally, as shown in~\cref{tab:diff_resource}, the vector set model trains about 2.5 times faster and uses half the VRAM, benefiting from three times fewer tokens with a smaller $V$.

\begin{table}
\centering
\caption{We average times on a RTX 2080 Ti over 5 runs for generating $21\times21$ chunks using RePaint with 5 resampling steps and our explicit outpainting method with the vecset diffusion model, and also report the triplane diffusion model's generation time.}
\vspace{-1em}
\resizebox{\linewidth}{!}
{
\begin{tabular}{@{}rrrrr@{}}
\toprule
Method & Outpaint & Scene Size & Emb Gen Time (s) & Occ Decode Time (s) \\
\midrule
vecset & RePaint & $21\times21$ & 1022.20 & 60.28 \\
triplane & explicit & $21\times21$ & 221.68 & 30.84 \\
vecset & explicit & $21\times21$ & 215.92 & 87.76 \\
\bottomrule
\end{tabular}
}
\label{tab:diff_inference_speed}
\end{table}

\mypara{Outpainting.} We compare against the inpainting method RePaint~\cite{lugmayr2022repaint}, widely used in prior works~\cite{wu2024blockfusion, meng2024lt3sd, lee2024semcity} for unbounded outpainting. Scenes are generated in the same order and with the same overlaps as our outpainting model. However, in RePaint, the mask and conditional embeddings are set to zero, with existing chunk conditions incorporated through its iterative re-sampling process. RePaint results for resampling steps $r=1$ and $r=5$ are shown in~\cref{fig:repaint}. Without resampling ($r=1$), inter-chunk connections lack coherence, and the process sometimes collapses into broken chunks, likely due to conditioning on noisy diffusion-generated chunks. With $r=5$, coherence improves, but the RePaint baseline produces less diverse scenes and occasional collapses (see supplement). Furthermore, re-sampling is significantly slower than our outpainting method (\cref{tab:diff_inference_speed}), as inference time scales with the number of resampling steps, whereas our method does not require resampling. Note our outpainting model is parallel and can be potentially adapted into other methods~\cite{wu2024blockfusion, meng2024lt3sd} for faster generation.

Our outpainting method is not perfect, and some chunks may still be poorly connected and contain noisy artifacts. In~\cref{fig:init-results}, the right orange box shows coherently connected chunks for our vector set diffusion model, while the left-side orange box highlights issues such as misaligned bridges or non-continuous chunks with visible gaps and seams.

\subsection{Multiple Scenes}

In~\cref{sec:exp_single_scene}, we demonstrated the efficiency of vector sets, allowing handling of scene chunks with varying heights. As well as our outpainting model for fast large scale outdoor scene generation. However, training on a single scene leads to generated sub-scenes resembling the training data, despite the model’s ability to generate some novel compositions (see supplement). To assess its generalization ability and validate our curation process for NuiScene43, we extend training to three additional scenes using the vector set backbone.

\cref{fig:multiscene} (b) shows a scene generated by our model, with purple boxes indicating how it integrates elements (rural houses, aqueduct, skyscraper) from two distinct scenes. The results demonstrates our model's generalization ability, as such combinations are absent from training, and highlights the effectiveness of our curation process which unifies diverse scenes into a cohesive dataset, enabling the stitching of chunks from disjoint environments.

We further compare our approach with Trellis~\cite{xiang2024structured}, a state-of-the-art image-to-3D object generation method, using TRELLIS-image-large without decimation. \cref{fig:multiscene} (a) shows that  compared to our model, Trellis produces coarser geometric details, as its single latent representation for entire scenes limits fine details in large environments. In contrast, our model represents each chunk separately, enabling higher-detail and unbounded scene generation. This highlights a key limitation of current object generation methods, which struggle to scale effectively for unbounded scene generation. See more results in the supplement.

\section{Limitations}

Currently, we are constrained to training on a limited number of scenes due to the high storage costs incurred by offline pre-sampling of chunks. More efficient online chunk sampling could enable training on our larger NuiScene43 dataset and improve inter-chunk consistency, potentially addressing issues where the model occasionally ignores conditions and generates discontinuous or misaligned chunks.

Another limitation is the lack of control or conditioning in our generative model, as we do not have labels for individual chunks. Leveraging foundation models~\cite{kirillov2023segment, achiam2023gpt} for semi-automatic label generation of semantic maps or captions may be used for conditioning during training. A related challenge, shared by prior works~\cite{wu2024blockfusion, meng2024lt3sd, lee2024semcity, liu2024pyramid}, is the absence of global context. The model only considers a limited local region, hindering large-scale planning. Exploring a global model capable of planning over large regions to mimic human-like design, such as coherent city layouts and road networks, is a promising direction for future work.

Finally, with the growing availability of 3D datasets~\cite{deitke2023objaversexl}, we aim to bridge the gap between methods like WonderWorld~\cite{yu2024wonderworld}, which rely on 2D priors from T2I models for open-domain world generation. This mirrors the progression from DreamFusion~\cite{poole2022dreamfusion}, which distills T2I models for object-centric 3D generation, to models like TRELLIS that train directly on 3D data for faster and more consistent generation. While our work also leverages 3D supervision, the scale remains limited; as future work, we plan to expand our dataset further by semi-automating our curation pipeline. Nevertheless, we believe our dataset can serve as the basis of a benchmark for unbounded outdoor scene generation.

\section{Conclusion}

We introduce \textbf{NuiScene}, a method for generating expansive outdoor scenes across varied environments. Our model tackles key challenges such as large height variations and efficient compression of scene chunks using vector sets. In addition, our explicit outpainting model enables fast inference and improved coherence. To support this task, we curate \textbf{NuiScene43}, a dataset of moderate to large-scale scenes with varying styles, suitable for unified training.

\newline
\mypara{Acknowledgements.}
This work was funded by a CIFAR AI Chair, an NSERC Discovery grant, and a CFI/BCKDF JELF grant. We thank Jiayi Liu, and Xingguang Yan for helpful suggestions on improving the paper.

{\small
\bibliographystyle{h_ieeenat_fullname.bst}
\bibliography{main}
}

\appendix
\renewcommand{\thesection}{\Alph{section}}

\twocolumn[{
    \renewcommand\twocolumn[1][]{#1}
    \maketitlesupplementary
    \begin{center}
\includegraphics[width=0.85\textwidth]{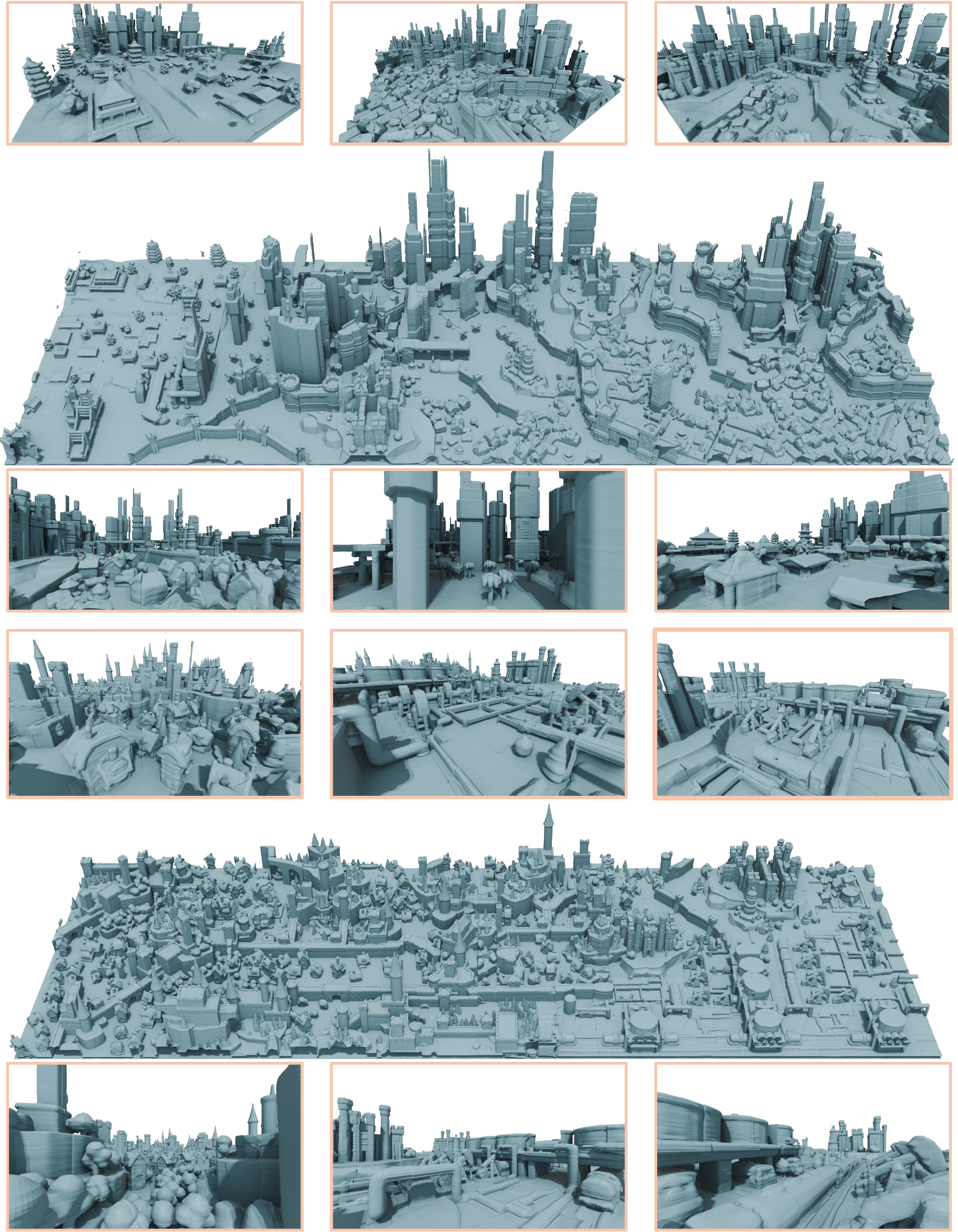}
\captionof{figure}{Additional generation results for our vector set diffusion model trained on 13 scenes. We show zoomed-in aerial shots on the top with street level views on the bottom respectively for the two scenes. The scenes shown here are of size $16\times46$.}
\label{fig:13_scene_large}
\end{center}

}]

\begin{figure*}
\centering
\includegraphics[width=\linewidth]{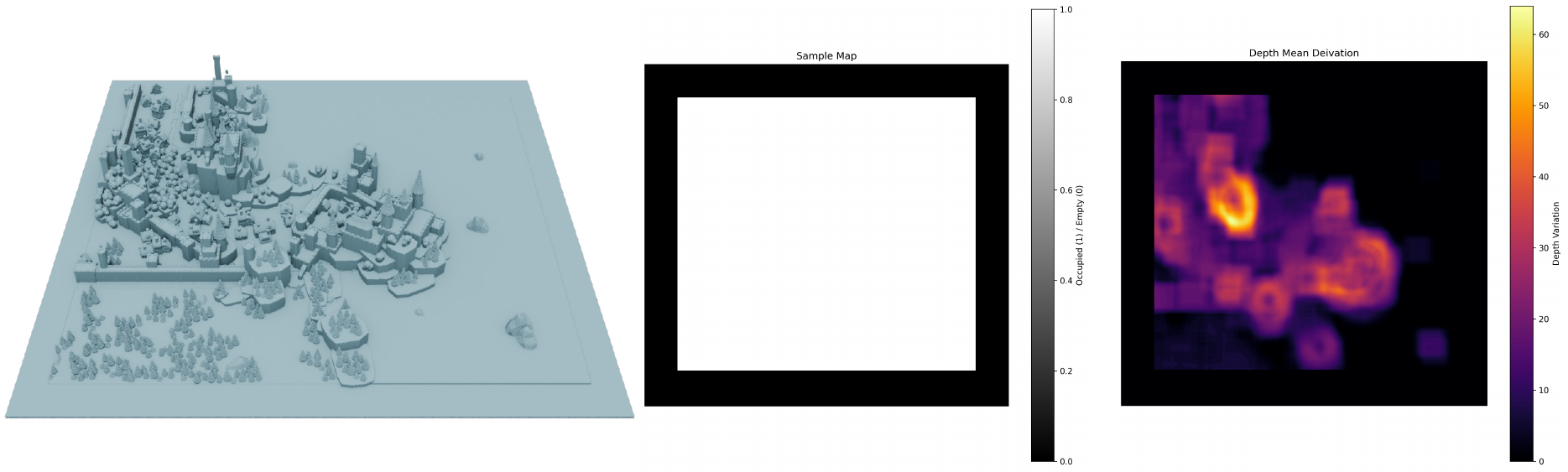}
\caption{Sample and depth mean deviation maps are calculated for sampling chunks from scenes.}
\label{fig:sample_map}
\end{figure*}

\begin{figure*}
\centering
\includegraphics[width=0.95\linewidth]{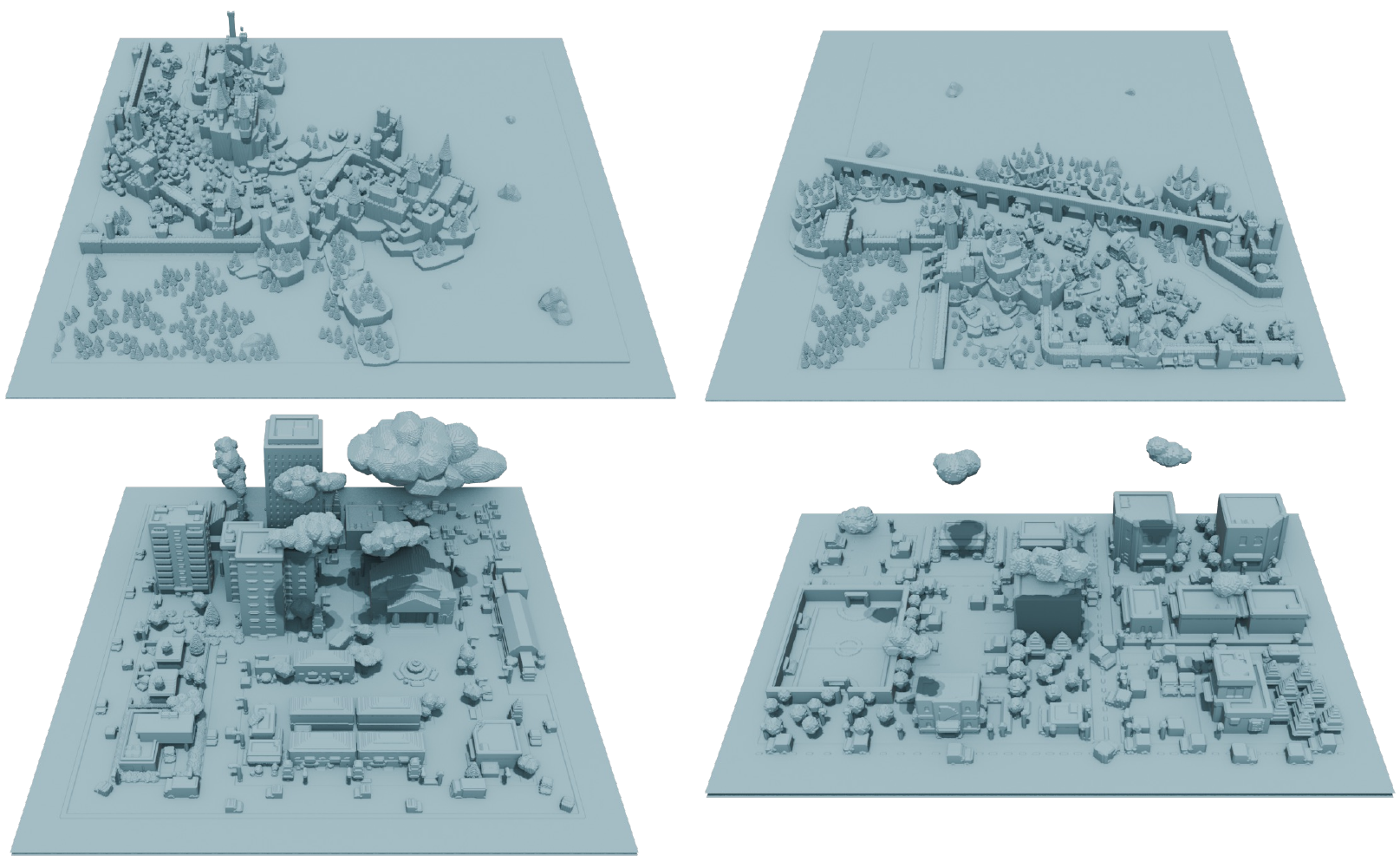}
\caption{Three additional scenes used to train our 4-scene model. The top two sub-scenes are split from a large Objaverse scene for occupancy calculation. All scenes have fixed ground geometries, and their meshes are extracted via marching cubes on the occupancy grid.}
\label{fig:nuiscene_4}
\end{figure*}

\section{NuiScene43 Dataset}
\label{sec:nuiscene43}

We filter scenes from Objaverse to select 43 high quality moderately to large sized scenes (\ref{sec:data-filtering}). To ensure a unified scale between scenes, we annotate scenes with scales relative to each other, aligning them under a uniform scale (\ref{sec:data-labeling}). Finally, we preprocess the raw Objaverse mesh files for training by sampling point clouds, adjusting ground geometry, and converting them to occupancies (\ref{sec:data-geometry}). Furthermore, we describe the sampling maps used to sample quad chunks from scenes (\ref{sec:chunk_sampling}) as well as some dataset statistics (\ref{sec:nuiscene_stat}). Please see our dataset page \url{https://3dlg-hcvc.github.io/NuiScene43-Dataset} for visualizations of all 43 scenes in NuiScene43.

\subsection{Filtering}
\label{sec:data-filtering}

We begin by filtering Objaverse~\cite{deitke2023objaverse} using multi-view embeddings from 12 object frames using DuoduoCLIP~\cite{lee2024duoduo} . We use cosine similarity to query embeddings using text and images to retrieve scenes, then refine selection through manual labeling and neural network filtering trained on DuoduoCLIP embeddings, reducing 37k initial scenes to 2k. Since manually processing all scenes for scale labeling and ground alignment is impractical, we select 43 larger scenes for initial experiments.

\subsection{Scale Labeling}
\label{sec:data-labeling}

To establish a uniform scale across scenes, we first normalize all scenes to $[-1,1]$. We then randomly select one scene as the anchor, assigning it a scale of 1. Each remaining scene is compared with the anchor, and its scale is adjusted visually to align elements such as trees or buildings from multiple angles, to ensure a visually coherent relative size. Since these scenes are artist-created, their proportions may prioritize aesthetic appeal over real-world accuracy. As a result, there is no definitive correct scale. Instead, we approximate a visually consistent scaling and record the assigned scale for each scene, with the anchor scene remaining fixed at scale 1.

\subsection{Geometry Processing}
\label{sec:data-geometry}

\mypara{Point Cloud and Occupancy.} All scene meshes are first converted into SDFs using the same method as~\citet{wang2022dual}, re-implemented in Taichi~\cite{hu2019taichi} for faster conversion. The scales obtained in~\ref{sec:data-labeling} are used to adjust the voxel grid resolution for SDF conversion to enforce the unified scale across scenes. We then convert to occupancy by thresholding SDF values. Finally, multiple iterations of flood filling are applied to fill in holes in the scene. Point clouds are sampled from the meshes after applying the Marching Cubes algorithm to the occupancy of each scene.

\mypara{Ground Fixing.} We observed that the filtered scenes had various ground geometry, ranging from flat planes to thick volumes. To deal with this, we identify the lowest ground level in each scene and enforce a uniform ground thickness of 5 voxels below this level in the occupancy grid, ensuring consistency in ground geometry across scenes.

\subsection{Chunk Sampling}
\label{sec:chunk_sampling}

To sample training chunks from scenes, we first compute an alpha map from the top-down view. Next, we apply a convolution with all one kernel weights over the entire scene using a kernel size of $(100, 100)$, the size of a quad chunk along the $x$ and $z$ axis. The resulting convolution map is then thresholded at a value of $100\times100$ to determine valid sampling locations. This ensures that all sampled chunks contain occupied regions, avoiding holes or boundary areas in the scenes.

Next, we compute a depth variation map. For each pixel, we calculate the mean depth within the kernel window and subtract the pixel’s depth value, taking the absolute value to obtain depth mean deviations. To avoid sampling overly flat regions, we filter out sample locations where the depth variation is below 2.5. An illustration of the sample maps is provided in~\cref{fig:sample_map}.

Finally, we apply farthest point sampling (FPS) to select quad chunks from all valid sampling locations. This helps minimize excessive overlap between chunks while ensuring maximum scene coverage for training.

\begin{table}
\centering
\caption{NuiScene43 statistics.}
\vspace{-1em}
\resizebox{\linewidth}{!}
{
\begin{tabular}{@{}lrr@{}}
\toprule
Category & \# Scenes & \# Sampleable Quad Chunks \\
\midrule
Rural/Medieval & 16 & 6.2 M \\
Low Poly City & 19 & 4.9 M\\
Japanese Buildings & 4 & 2.9 M \\
Other & 4 & 1.2 M \\
\bottomrule
\end{tabular}
}
\label{tab:nuiscene_stat}
\end{table}

\subsection{NuiScene43 Statistics}
\label{sec:nuiscene_stat}

We present statistics of NuiScene43 in~\cref{tab:nuiscene_stat}, including the number of scenes in each category and the total number of quad chunks that can be sampled. The chunk count is derived by summing the sampling map discussed earlier, following depth variation thresholding. Note that these chunks may overlap, as we consider all valid $x$ and $z$ coordinates.

\begin{algorithm*}
\caption{Raster Scan Order Scene Generation}
\label{alg:raster_generation}
\begin{algorithmic}[1]
\Require Number of rows $I$, number of columns $J$, trained diffusion model $\epsilon_\theta$
\State Initialize scene latent grid $\bm{Z} \in \mathbb{R}^{I \times J \times V \times c}$
\For{$i = 0$ to $I-2$} \Comment{Iterate row-wise}
    \For{$j = 0$ to $J-2$} \Comment{Iterate column-wise}
        \If{$i = 0$ and $j = 0$} \Comment{First chunk (Full)}
            \State $M \gets \{0, 0, 0, 0\}$
            \State $Z_\text{cond} \gets \{\mathbf{0}, \mathbf{0}, \mathbf{0}, \mathbf{0}\}$
        \ElsIf{$i = 0$} \Comment{First row (Left-Right)}
            \State $M \gets \{1, 0, 1, 0\}$
            \State $Z_\text{cond} \gets \{\bm{Z}_{i, j}, \mathbf{0}, \bm{Z}_{i+1, j}, \mathbf{0}\}$
        \ElsIf{$j = 0$} \Comment{First column (Top-Down)}
            \State $M \gets \{1, 1, 0, 0\}$
            \State $Z_\text{cond} \gets \{\bm{Z}_{i, j}, \bm{Z}_{i, j+1}, \mathbf{0}, \mathbf{0}\}$
        \Else \Comment{All other cases (Diagonal)}
            \State $M \gets \{1, 1, 1, 0\}$
            \State $Z_\text{cond} \gets \{\bm{Z}_{i, j}, \bm{Z}_{i, j+1}, \bm{Z}_{i+1, j}, \mathbf{0}\}$
        \EndIf
        \State $\bm{X}_T \sim \mathcal{N}(0, \mathbf{I})$ \Comment{Sample Gaussian noise}
        \State $\bm{C} \gets M \oplus Z_{\text{cond}} \oplus \text{PE}$ \Comment{Conditioning input}
        \State $\{\bm{z}_0, \bm{z}_1, \bm{z}_2, \bm{z}_3\} \gets \text{Denoise}(\bm{X}_T \oplus \bm{C}, \epsilon_\theta)$ \Comment{Denoise 2×2 quad chunk}
        \State $\bm{Z}_{i, j}, \bm{Z}_{i, j+1}, \bm{Z}_{i+1, j}, \bm{Z}_{i+1, j+1} \gets \bm{z}_0, \bm{z}_1, \bm{z}_2, \bm{z}_3$ \Comment{Write to scene latent grid}
    \EndFor
\EndFor
\State \Return $\bm{Z}$
\end{algorithmic}
\end{algorithm*}

\section{Implementation Details}

\subsection{4-scene Training}

We add an additional 3 scenes along with the original single scene for training the 4-scene model in the main paper. The 3 additional scenes are shown in~\cref{fig:nuiscene_4}. 

\subsection{Raster Scan Order Generation}

We show the algorithm of our raster scan order generation during inference utilizing our diffusion model trained with 4 different configurations in~\cref{alg:raster_generation}. The algorithm outlines how the 4 different masking configurations are used to condition on existing chunks for continuous and unbounded generation. In line 19 of the algorithm we denoise for 50 steps using the DDPM scheduler. After we obtain a large grid of embeddings, the decoder is used to decode occupanices followed by marching cubes to get the final mesh.

It can be observed that generation of the next row can begin once a single quad chunk has been generated in the current row. This allows for parallelization along the scene’s anti-diagonal, greatly accelerating sampling for large scenes. However, memory usage becomes inconsistent and varies with scene size. For consistency and simplicity, we use the sequential raster scan order (\cref{alg:raster_generation}) in the paper. Please refer to our \href{https://github.com/3dlg-hcvc/NuiScene}{implementation} for more details.

\subsection{Details on Scene Texturing}
\label{supp-texture}

\begin{figure*}[!t]
  \begin{subfigure}{1.0\linewidth}
      \includegraphics[width=\textwidth]{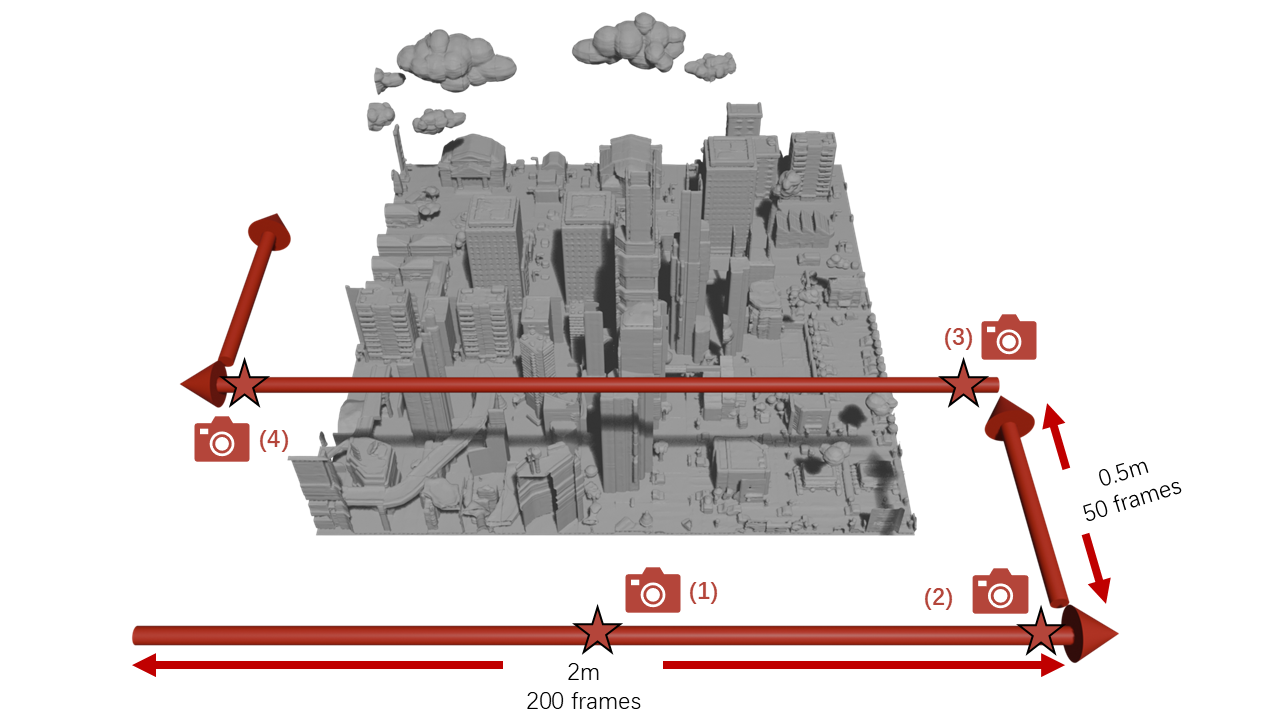}
      \caption{An illustration of our choice for the camera trajectory, indicated by red arrows.}
      \label{fig:texture-overview}
  \end{subfigure}
  
  \begin{subfigure}{0.24\linewidth}
      \includegraphics[width=\textwidth]{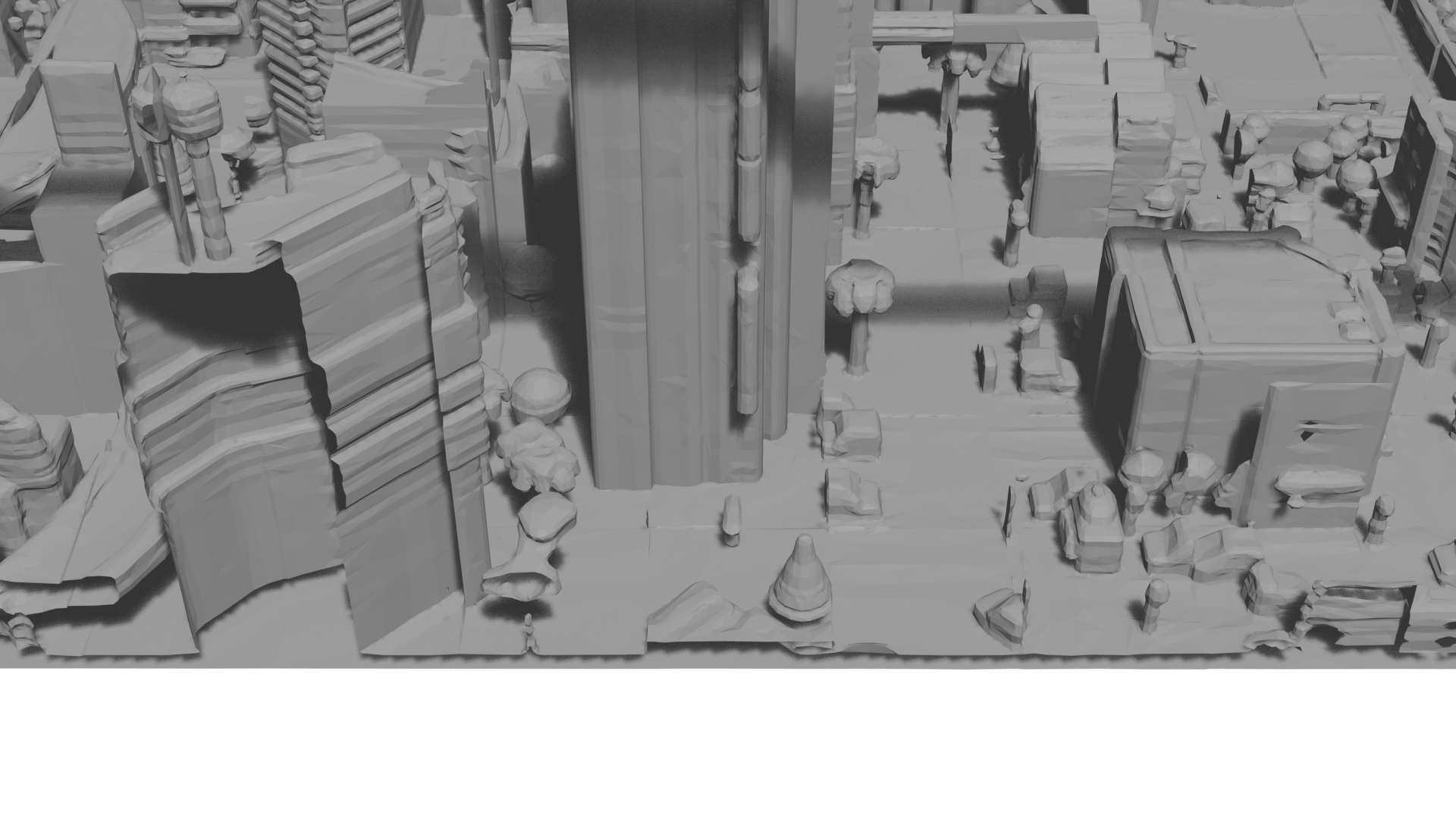}
      \caption{Image rendered at position (1).} 
      \label{fig:texture-100}
  \end{subfigure} 
  \hfill
  \begin{subfigure}{0.24\linewidth} 
      \includegraphics[width=\textwidth]{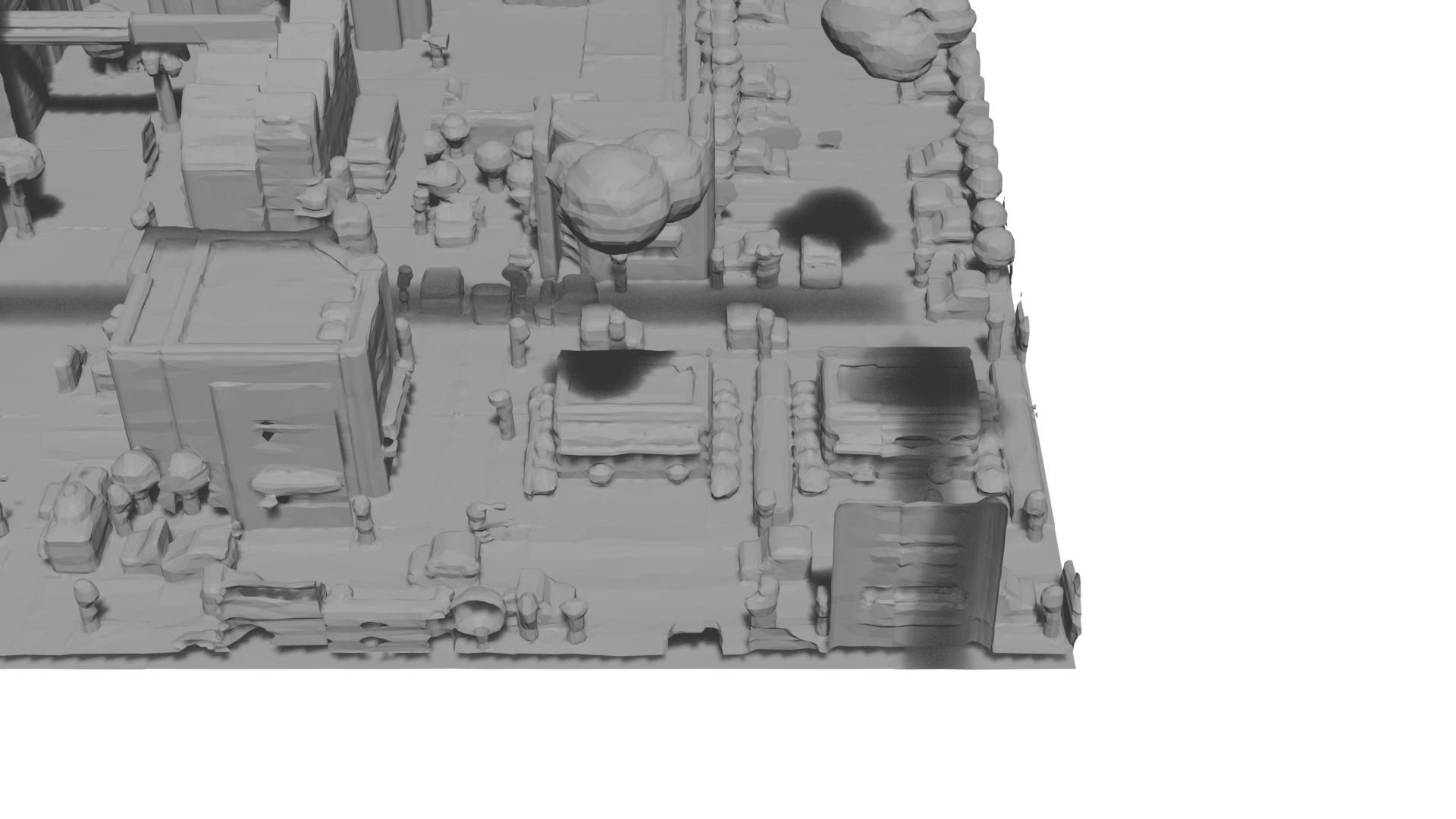} 
      \caption{Image rendered at position (2).} 
      \label{fig:texture-200}
  \end{subfigure}  
  \hfill
  \begin{subfigure}{0.24\linewidth} 
      \includegraphics[width=\textwidth]{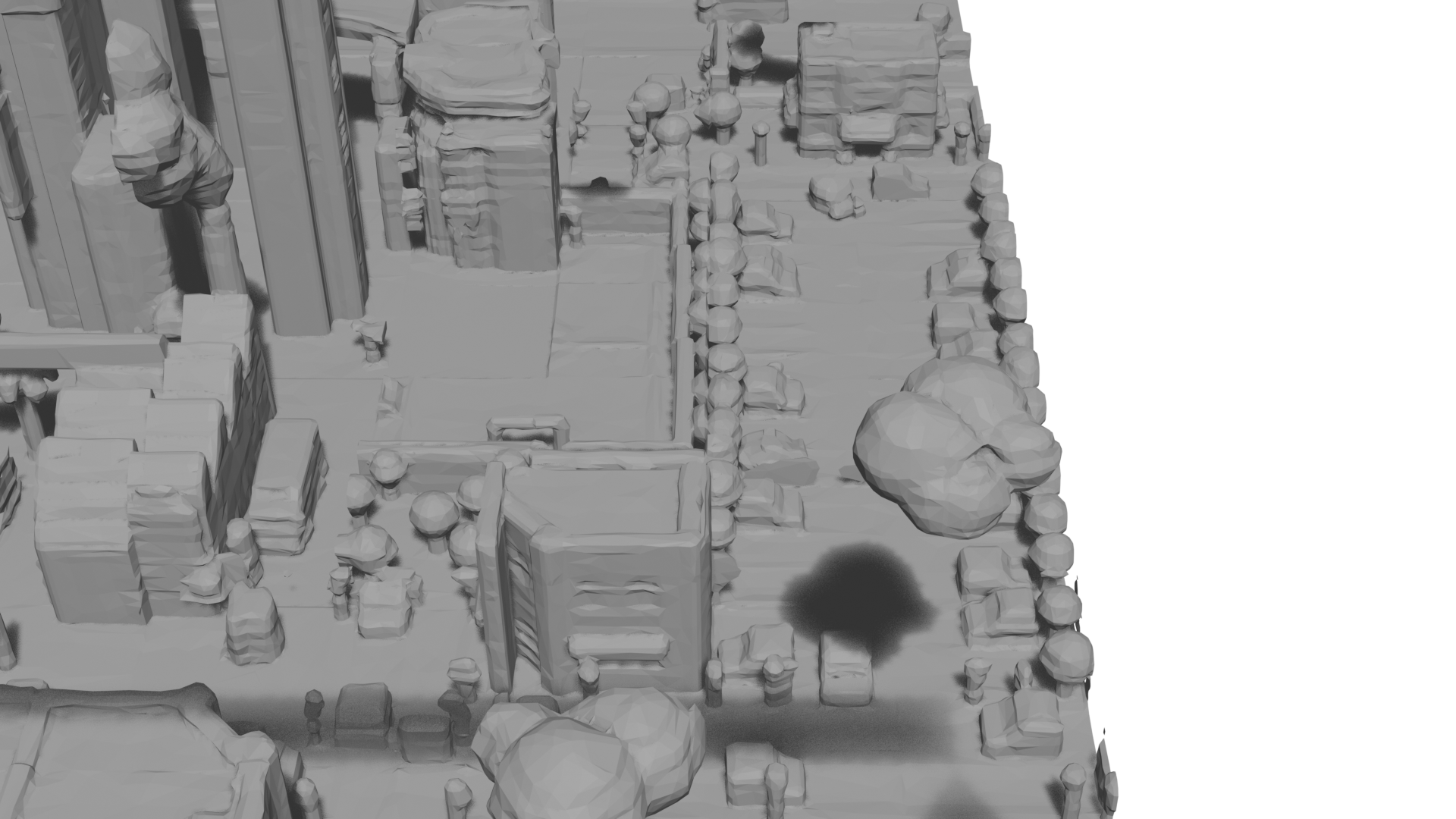} 
      \caption{Image rendered at position (3).}
      \label{fig:texture-250}
  \end{subfigure}
  \hfill
  \begin{subfigure}{0.24\linewidth} 
      \includegraphics[width=\textwidth]{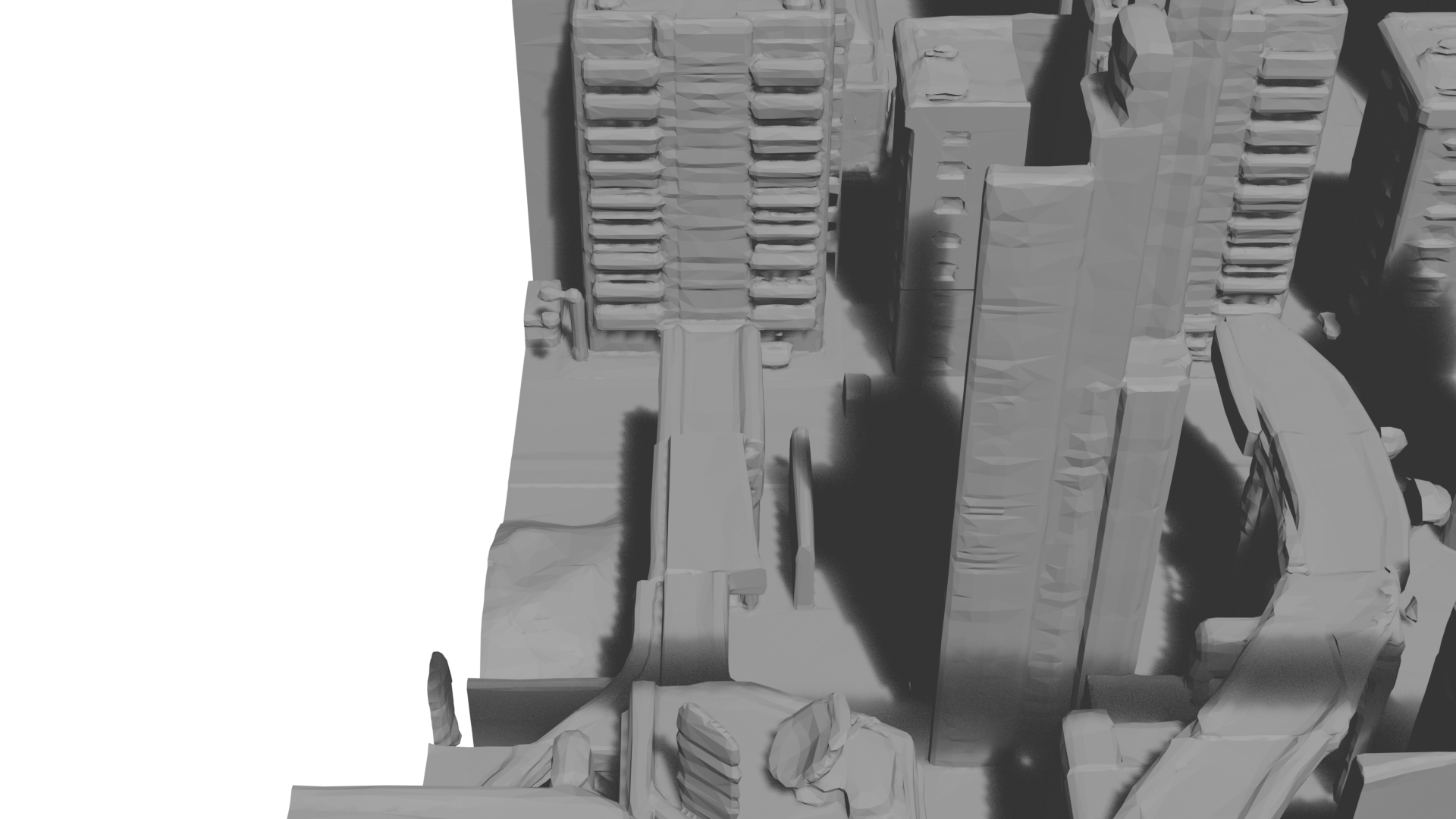} 
      \caption{Image rendered at position (4).}
      \label{fig:texture-450}
  \end{subfigure}
  
  \caption{An overview of our choice of the camera trajectory in Blender and the four images respectively rendered at position (1), (2), (3) and (4). We adopt a snake-scan trajectory pattern allowing for a more comprehensive coverage of the entire scene. The long side of the trajectory spans 2 meters at a fixed number of 200 frames, and the shorter side spans 0.5 meters for 50 frames. Depending on the shape of the scene, the total number of frames ranges from 1.2k to 1.8k.}
  \label{fig:texture}
\end{figure*}

\begin{figure*}[!t]
  \begin{subfigure}{0.24\linewidth}
      \includegraphics[width=\textwidth]{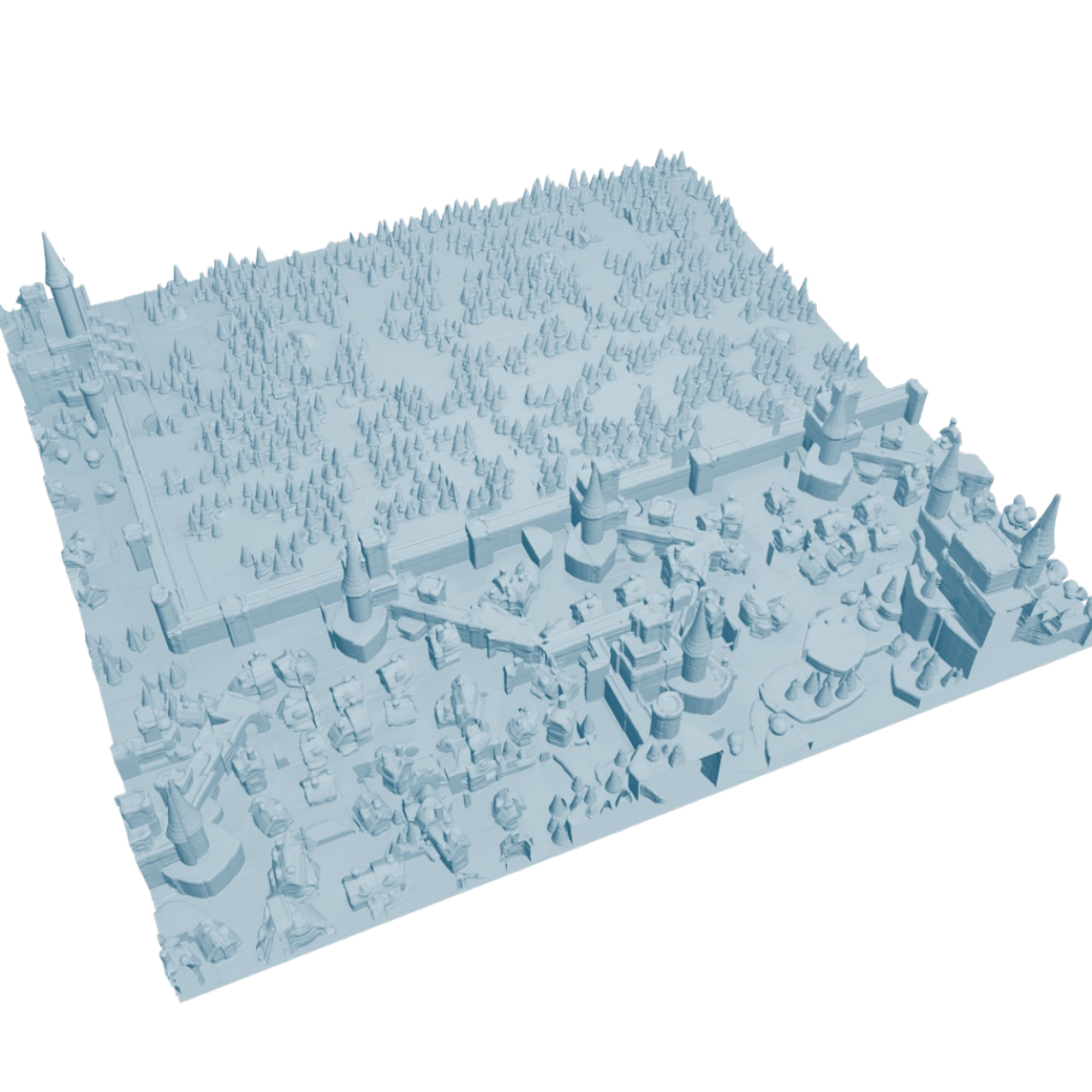}
      \caption{Generated scene (1).} 
  \end{subfigure} 
  \hfill
  \begin{subfigure}{0.24\linewidth} 
      \includegraphics[width=\textwidth]{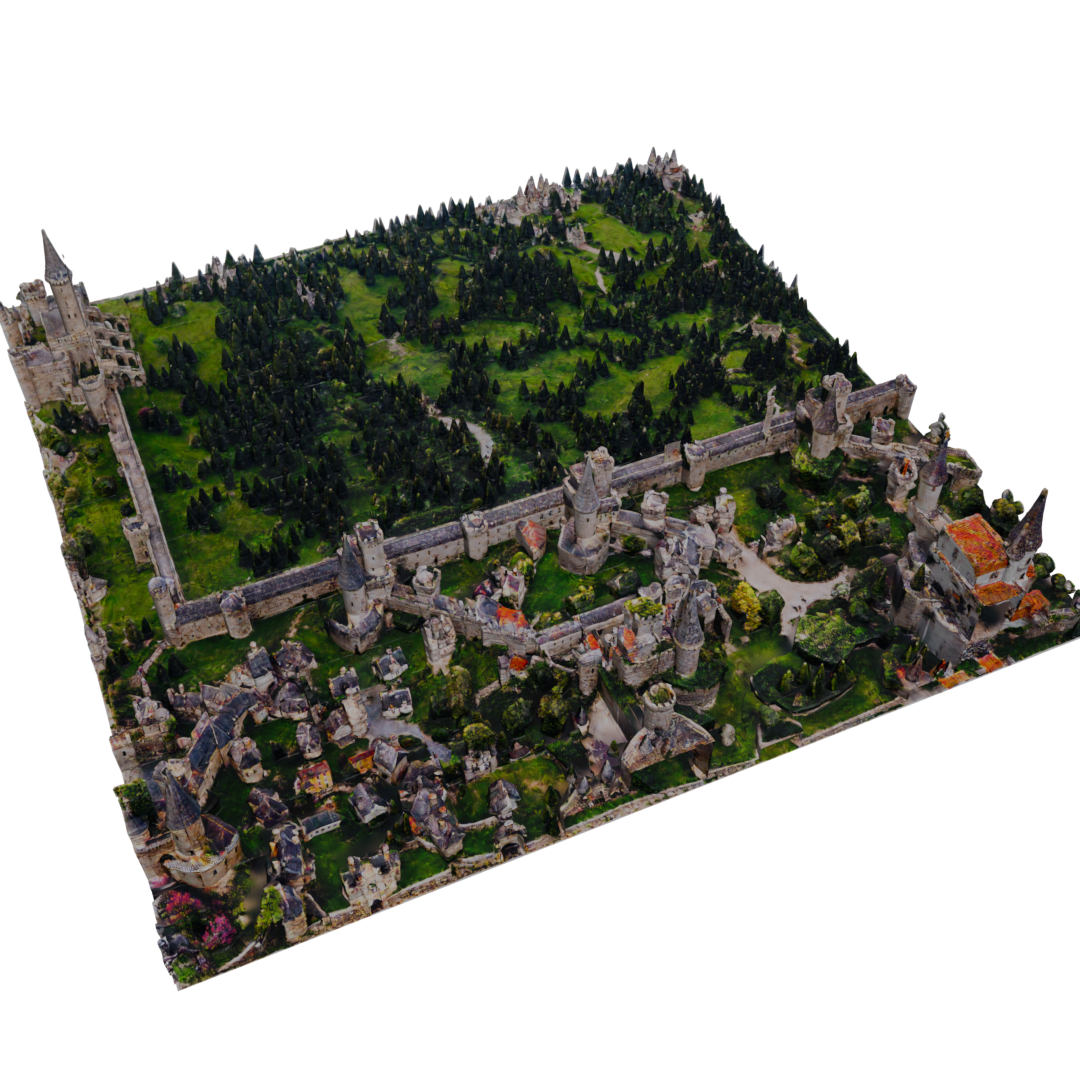} 
      \caption{Textured scene (1).} 
  \end{subfigure}  
  \hfill
  \begin{subfigure}{0.24\linewidth} 
      \includegraphics[width=\textwidth]{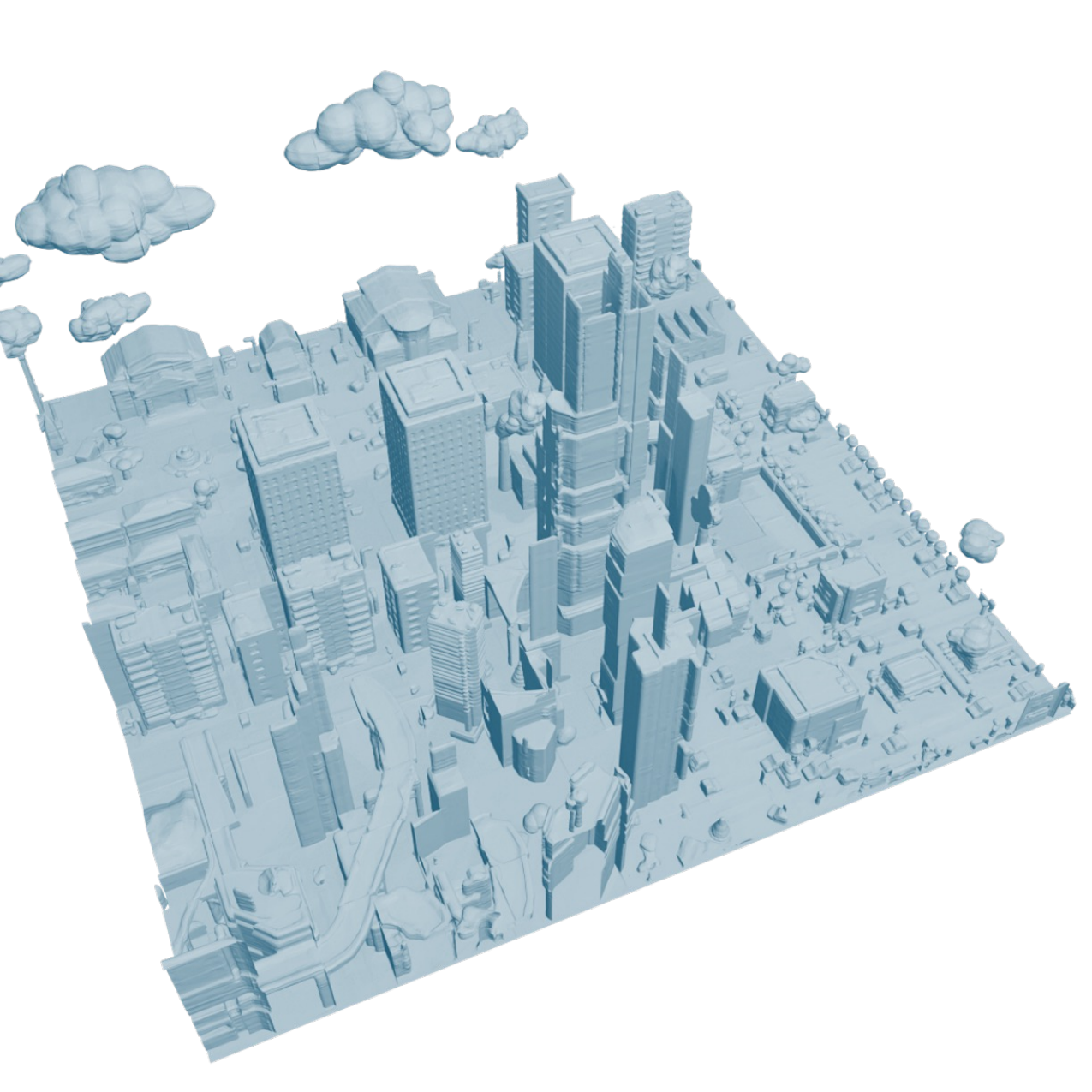} 
      \caption{Generated scene (2).}
  \end{subfigure}
  \hfill
  \begin{subfigure}{0.24\linewidth} 
      \includegraphics[width=\textwidth]{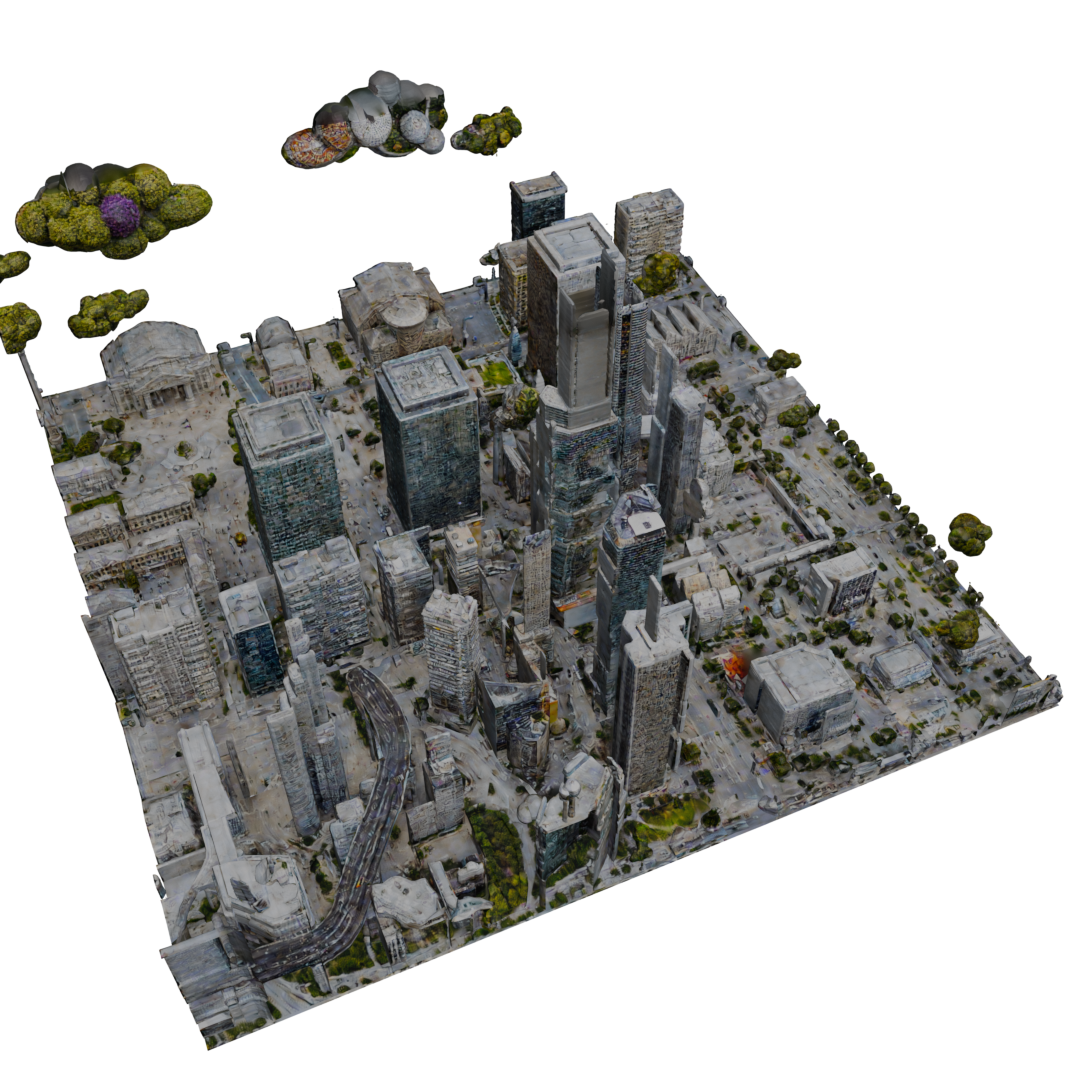} 
      \caption{Textured scene (2).}
  \end{subfigure}
  \begin{subfigure}{0.24\linewidth}
      \includegraphics[width=\textwidth]{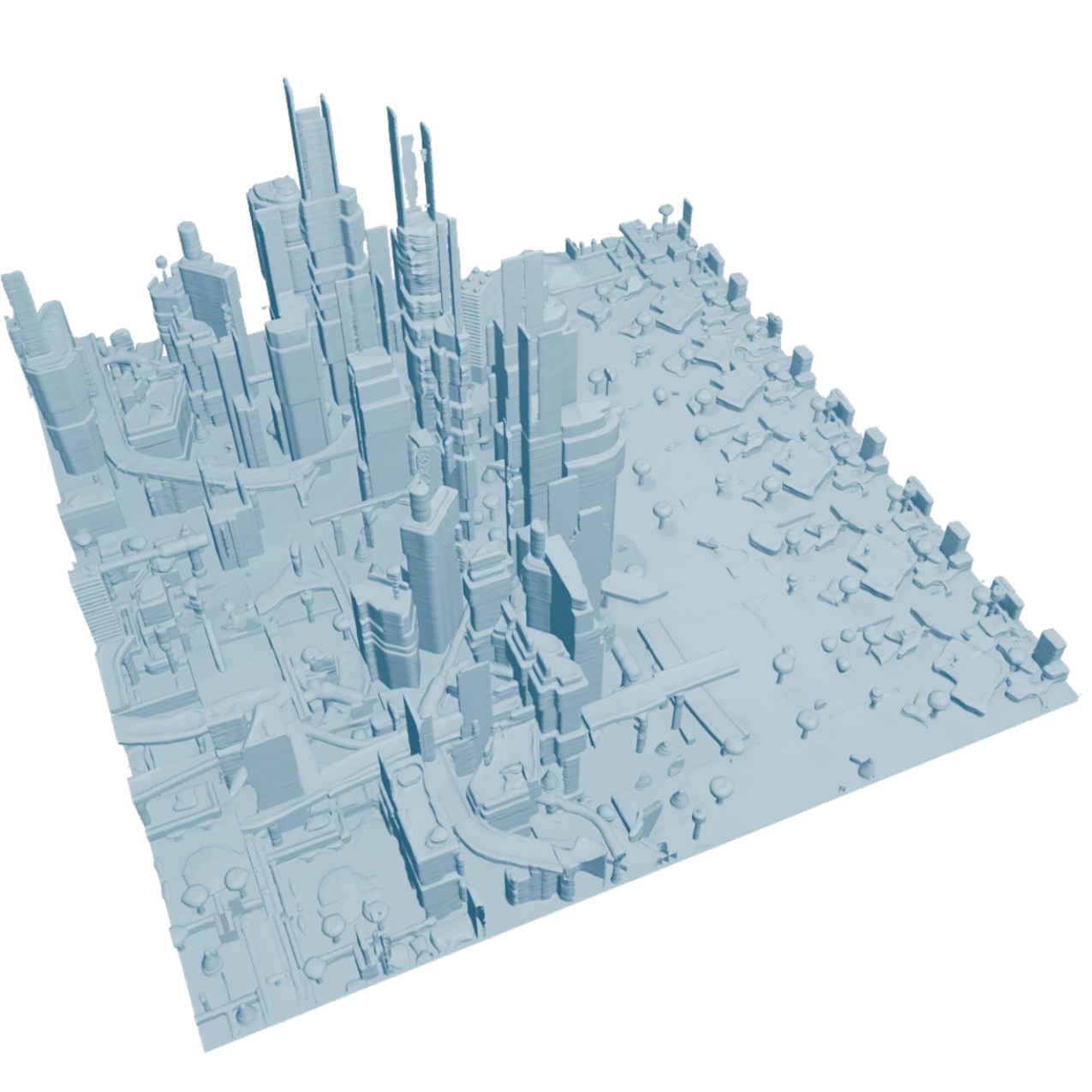}
      \caption{Generated scene (3).} 
  \end{subfigure} 
  \hfill
  \begin{subfigure}{0.24\linewidth} 
      \includegraphics[width=\textwidth]{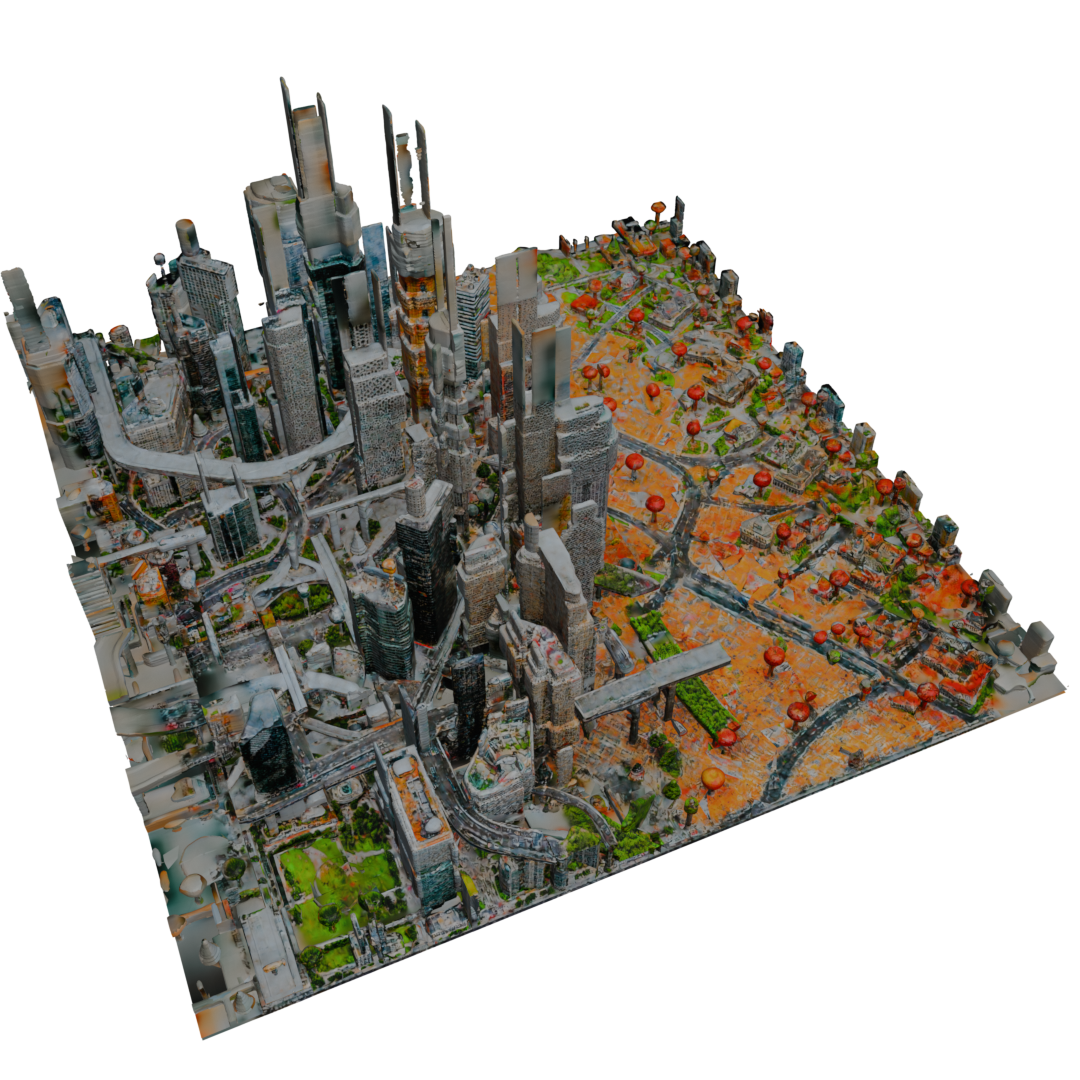} 
      \caption{Textured scene (3).} 
  \end{subfigure}  
  \hfill
  \begin{subfigure}{0.24\linewidth} 
      \includegraphics[width=\textwidth]{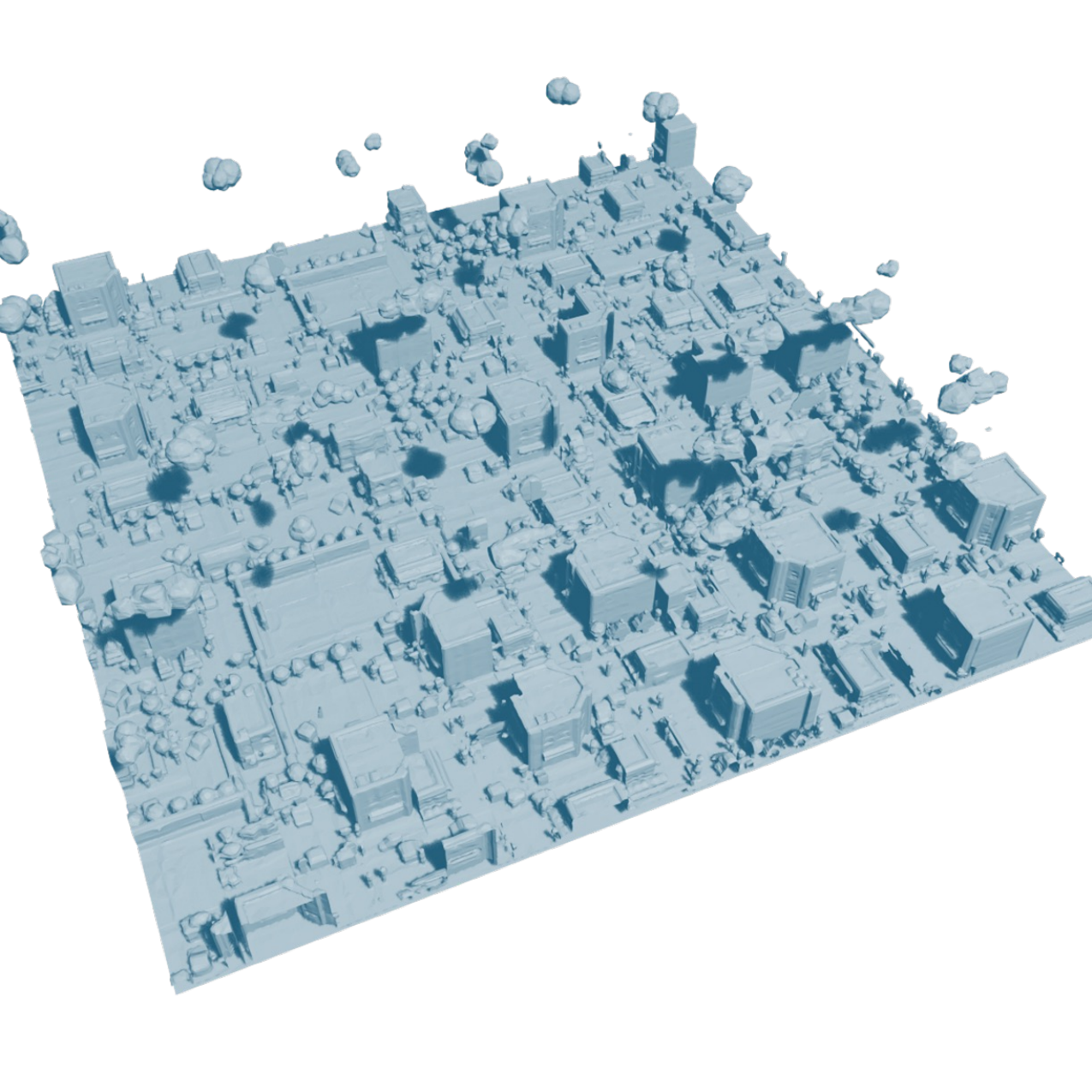} 
      \caption{Generated scene (4).}
  \end{subfigure}
  \hfill
  \begin{subfigure}{0.24\linewidth} 
      \includegraphics[width=\textwidth]{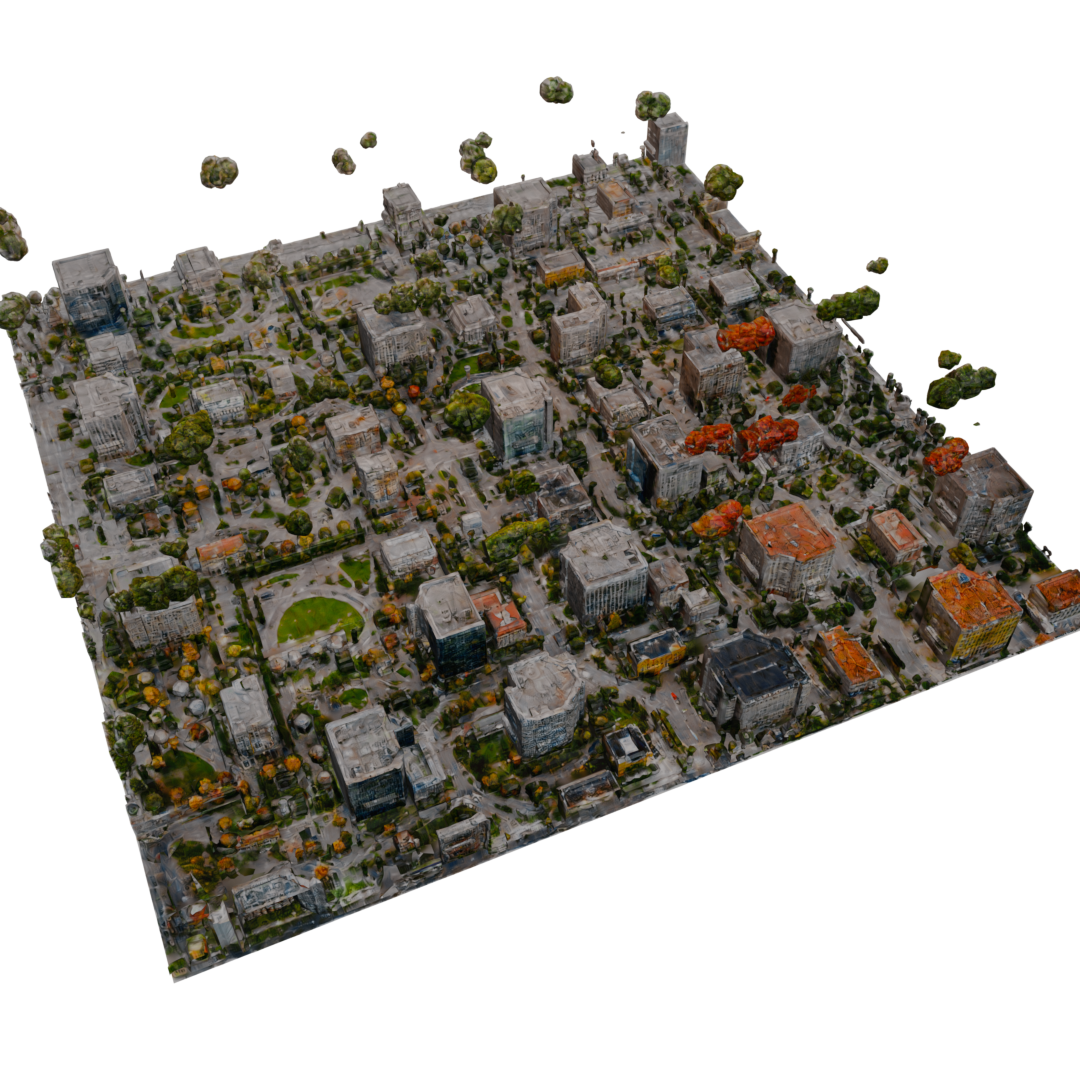} 
      \caption{Textured scene (4).}
  \end{subfigure}
  \caption{Four examples (1) to (4) of generated scenes and the textured counterparts. (1) is textured using prompt ``A beautiful town with buildings, castles and trees''. (2)(3) are textured using prompt ``A large city with buildings and surroundings''. (4) is textured using prompt ``A modern city with buildings and trees, clouds in the sky''.}
  \label{fig:texture-example}
\end{figure*}

SceneTex~\cite{chen2024scenetex} offers three camera modes\footnote{\url{https://github.com/daveredrum/SceneTex}} for rendering images and texture optimization, i.e., {\it Spherical Cameras}, the {\it Blender Cameras} and {\it BlenderProc Cameras}. For small indoor scenes like those from 3DFront~\cite{fu20213d}, spherical cameras are sufficient to capture the fine-grained details of the scene and produce high-quality textures. However, for large outdoor scenes, a predefined spherical camera trajectory often fails to cover the entire scene comprehensively and misses important scene details. We therefore choose the {\it Blender Cameras} mode to manually keyframe the camera trajectory for each scene. This enables a finer control over capturing the scene details, leading to an improved texture.

More specifically, we first normalize each scene into a spatial range of $[-1,1]^3$ centered at the world origin. Then, we define a snake-scan-like camera trajectory to capture the entire scene at a specified frame rate, as shown in \cref{fig:texture}. Each frame will be rendered by SceneTex and all rendered images are sampled during training to optimize the scene-level texture. Note that to enable faster optimization, we use Blender's {\it Decimate Geometry} and {\it Merge By Distance} with a distance of 0.001m to compress the scene until its size is smaller than 20MB. This might degrade the geometry for large scenes, but it's only for the texturing purpose. We encourage readers to focus on the untextured geometry details of the generated scenes. \cref{fig:texture-example} shows more examples on raw geometries and the textured counterparts.

\section{Additional Results}

\subsection{Single Scene Results}

\begin{table}
\centering
\caption{Additional quantitative comparison of reconstruction across different VAE backbones for the single scene experiment. Here $\hat{h}$ indicates that the predicted height was used for the occupancy prediction and $\tilde{h}$ the ground truth height.}
\vspace{-1em}
\resizebox{\linewidth}{!}
{
\begin{tabular}{@{}llrrrrr@{}}
\toprule
Method & Output Res/S & IOU$\uparrow$ & CD $(\hat{h})\downarrow$ & F-Score $(\hat{h})\uparrow$ & CD $(\tilde{h})\downarrow$ & F-Score $(\tilde{h})\uparrow$\\
\midrule
\multirow{2}{*}{triplane} & $3\times64^2$/$6$ & 0.940 & 0.064 & 0.831 & 0.064 & 0.831 \\
& $3\times128^2$/$6$ & 0.982 & 0.185 & $\mathbf{0.879}$ & 0.058 & 0.858 \\
\midrule
vecset & - & $\mathbf{0.989}$ & $\mathbf{0.055}$ & 0.864 & $\mathbf{0.055}$ & $\mathbf{0.863}$ \\
\bottomrule
\end{tabular}
}
\label{tab:vae_recon_deconv5}
\end{table}

\mypara{Triplane $3\times128^2$.} \cref{tab:vae_recon_deconv5} presents results for the Triplane baseline with an output resolution of $3\times128^2$. We see that the IOU outperforms the lower-resolution triplane baseline and beats the vecset model on F-Score($\hat{h}$). However, we notice a large increase in CD($\hat{h}$). Upon inspecting chunks with larger CD values, when predicted heights are larger than the ground truth, the model produced floating artifacts. This may be attributed to the higher resolution, which may be more sensitive to spatial aliasing or quantization artifacts during feature querying. Adjusting the scale factor to $S = 8.5$ may be more appropriate for this resolution, but due to the high training cost, we did not retrain the model. These results highlight the sensitivity of spatially structured latents to scene bounds and hyperparameters like $S$.

\begin{table}
\centering
\caption{Comparison of VAE training resources for vector set vs triplane backbones with larger batch sizes. Training for all experiments was run on 4 L40S GPUs, total batch size and memory across 4 gpus are reported. The \# Latents is the size of the latent for the VAE backbone and Output Res indicates the triplane size after deconvolution.}
\vspace{-1em}
\resizebox{\linewidth}{!}
{
\begin{tabular}{@{}rrrrrr@{}}
\toprule
Method & BS & \# Latents & Output Res & Time (hr) & Mem. (GB)\\
\midrule
\multirow{2}{*}{triplane} & 192 & $3\times4^2$ & $3\times32^2$ & 18 & 166.0 \\
& 120 & $3\times4^2$ & $3\times64^2$ & 29.4 & 164.4 \\
\midrule
vecset & 144 & 16 & - & 21.6 & 170.5 \\
\bottomrule
\end{tabular}
}
\label{tab:vae_resource_large_bs}
\hspace{1cm}
\end{table}

\begin{table}
\centering
\caption{Quantitative comparison of reconstruction across different VAE backbones using larger batch sizes in~\cref{tab:vae_resource_large_bs}. Here $\hat{h}$ indicates that the predicted height was used for the occupancy prediction and $\tilde{h}$ the ground truth height.}
\vspace{-1em}
\resizebox{\linewidth}{!}
{
\begin{tabular}{@{}llrrrrr@{}}
\toprule
Method & Output Res/S & IOU$\uparrow$ & CD $(\hat{h})\downarrow$ & F-Score $(\hat{h})\uparrow$ & CD $(\tilde{h})\downarrow$ & F-Score $(\tilde{h})\uparrow$\\
\midrule
\multirow{2}{*}{triplane} & $3\times32^2$/$6$ & 0.727 & 0.171 & 0.508 & 0.170 & 0.503 \\
& $3\times64^2$/$6$ & 0.933 & 0.099 & 0.852 & 0.064 & 0.830 \\
\midrule
vecset & - & $\mathbf{0.962}$ & $\mathbf{0.072}$ & $\mathbf{0.890}$ & $\mathbf{0.057}$ & $\mathbf{0.858}$ \\
\bottomrule
\end{tabular}
}
\label{tab:vae_recon_bs}
\end{table}

\begin{figure}
\centering
\includegraphics[width=0.80\linewidth]{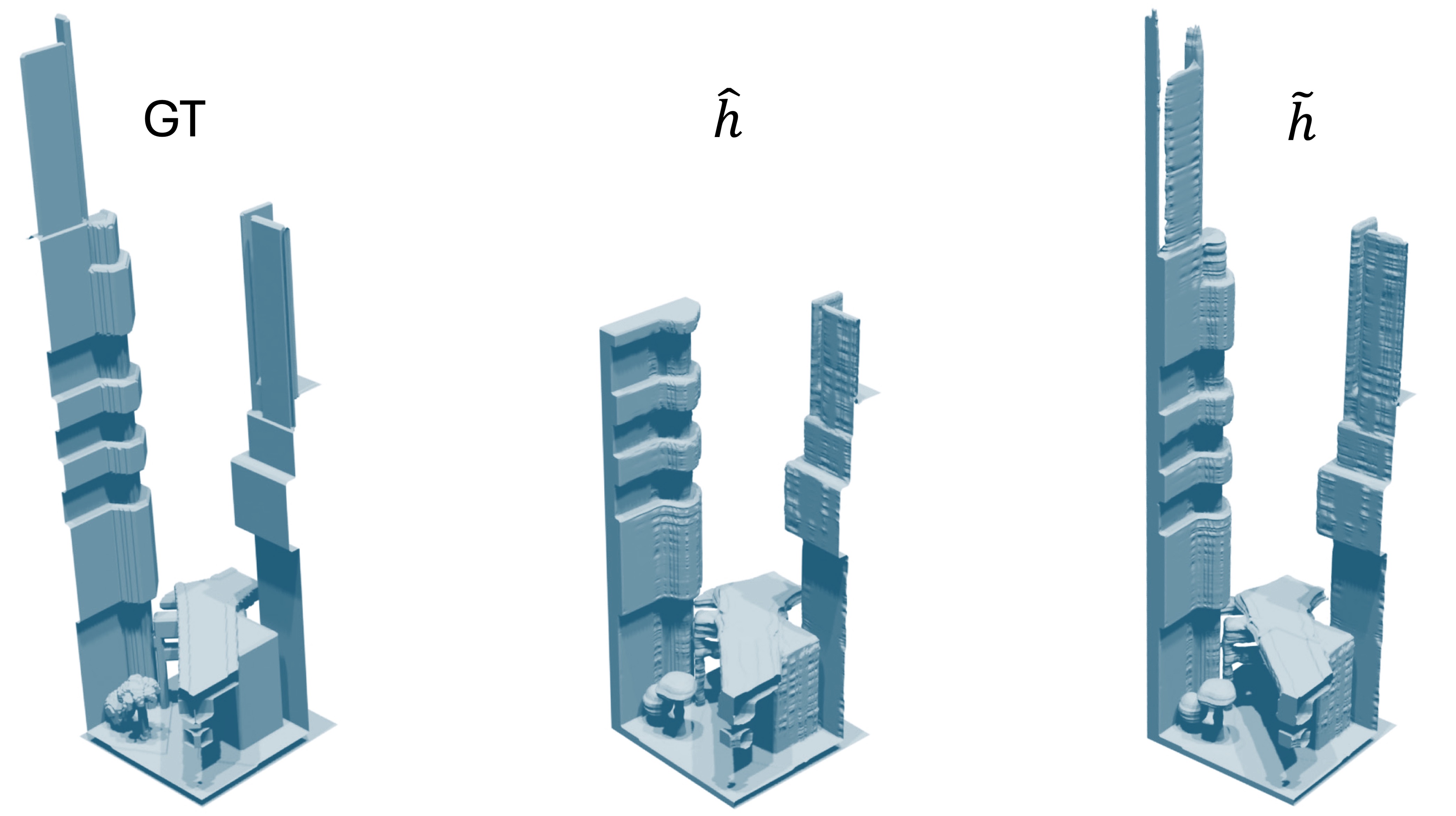}
\caption{We compare the reconstruction of the vector set model with larger batch size (144) when using the predicted height $\hat{h}$ of the model and using the ground truth height $\tilde{h}$.}
\label{fig:height_pred}
\end{figure}

\begin{figure*}
\centering
\includegraphics[width=0.95\linewidth]{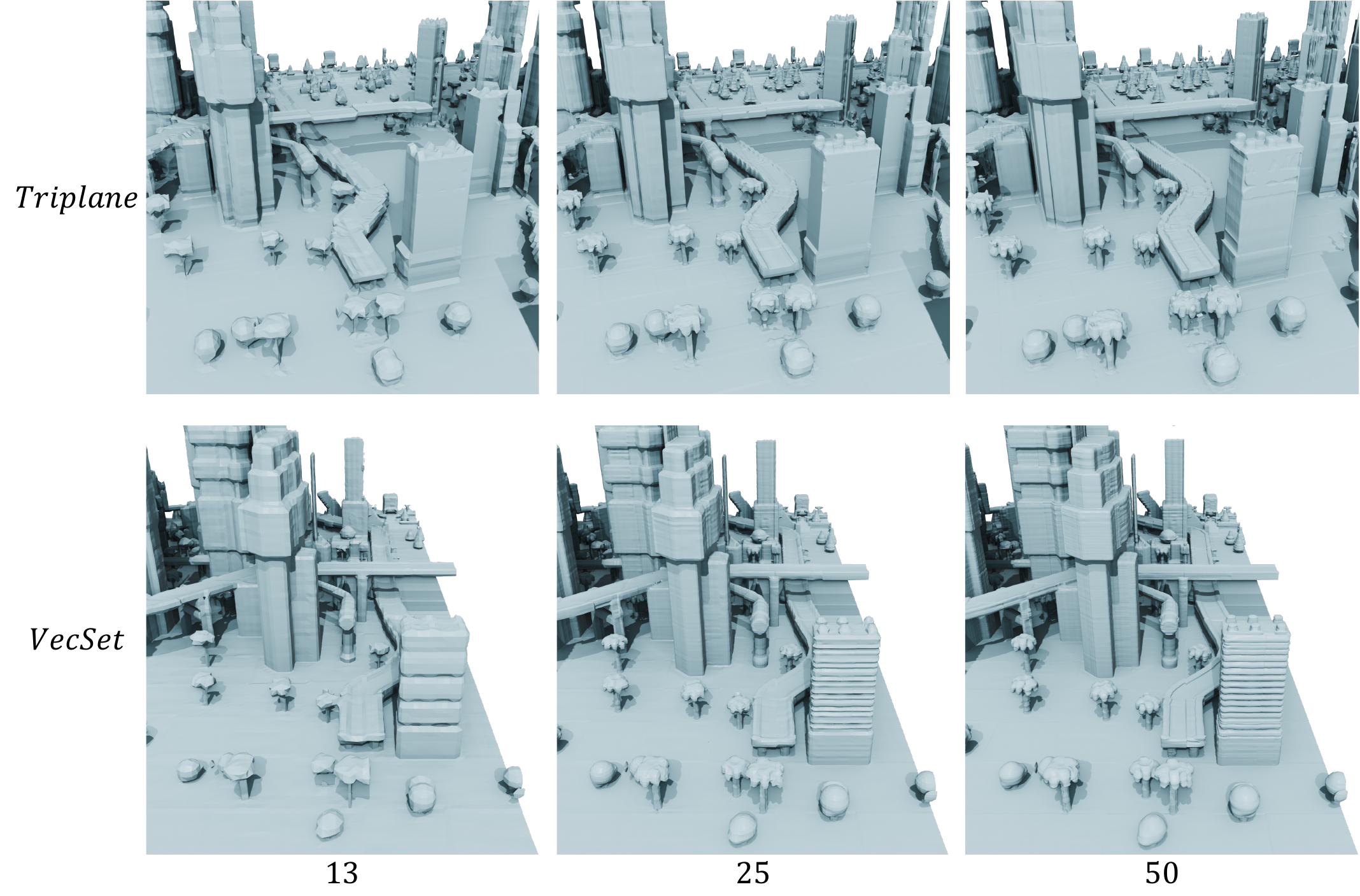}
\caption{Here we show the results of using different occupancy grid resolutions for prediction during inference for marching cubes. The numbers on the bottom indicate the chunk size along the $x$ and $z$ axis. $50$ is the original chunk size used for training.}
\label{fig:lod}
\end{figure*}

\begin{figure*}
\centering
\includegraphics[width=0.95\linewidth]{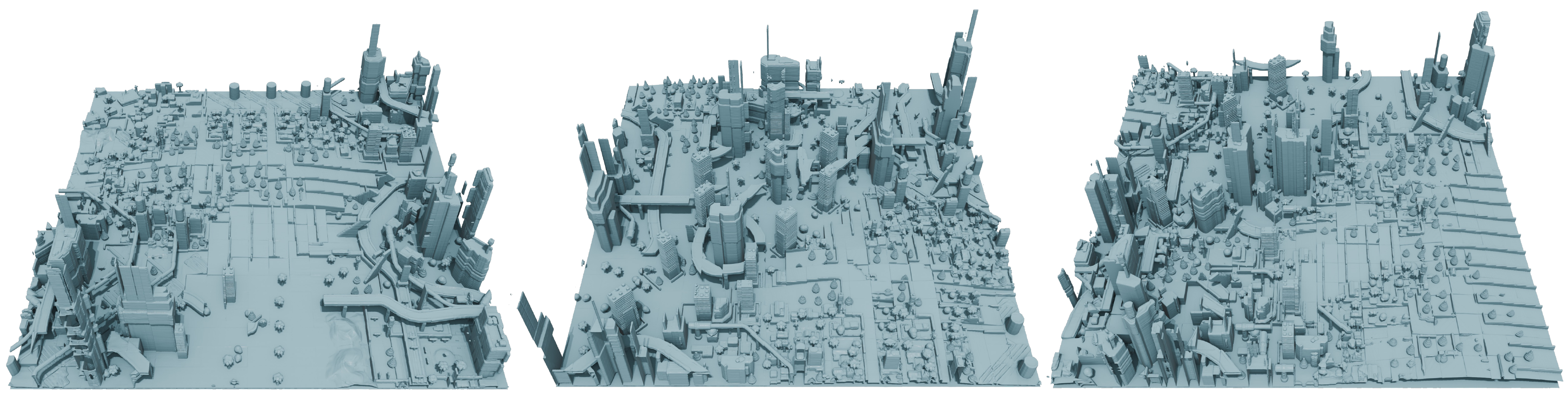}
\caption{Additional results for our vector set diffusion model trained on a single scene. The scenes shown here are of size $21\times21$.}
\label{fig:more_3b61}
\end{figure*}

\begin{figure*}
\centering
\includegraphics[width=0.95\linewidth]{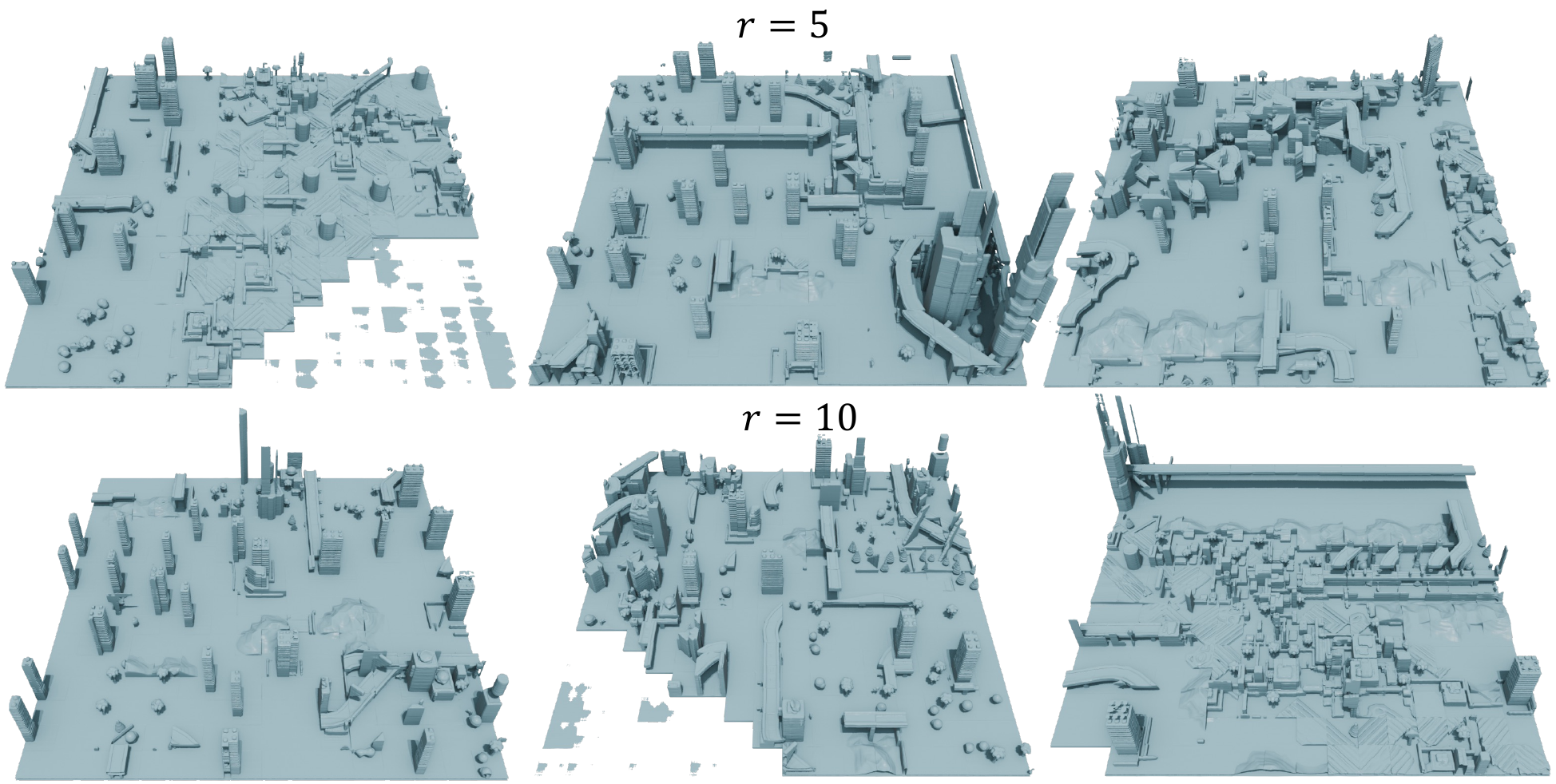}
\caption{Additional RePaint results using our vector set diffusion model trained on a single scene. The scenes shown here are of size $16\times16$. We show results for resampling steps $r=5$ and $r=10$. Compared to our outpainting model, RePaint struggles with inter-chunk coherence and sometimes collapses, producing broken chunks in larger scenes.}
\label{fig:more_3b61_repaint}
\end{figure*}

\mypara{Larger Batch Size.} We initially experimented with larger batch sizes across all configurations, using 4 GPUs, as shown in~\cref{tab:vae_resource_large_bs} to accelerate training. For the vecset model, this reduced training time to 21.6 hours from 36.1 hours on 2 GPUs (main paper configuration). However, it led to overfitting, as seen in~\cref{tab:vae_recon_bs} which also affected height prediction accuracy as demonstrated in~\cref{fig:height_pred} where the model under-predicted the height, leaving parts of the chunk unreconstructed. In contrast, triplane backbones were less affected by overfitting. We hypothesize this is due to their spatial locality where query points retrieve features from fixed triplane positions, whereas the vecset backbone allows queries to aggregate from all feature vectors, offering greater capacity but also more prune to overfitting. Nevertheless, this shows the limit of the one scene training.

\mypara{Evaluation Metrics.} Following 3DShape2Vecset~\cite{zhang20233dshape2vecset}, we report IOU, Chamfer Distance (CD), and F-Score for VAE reconstruction. However, we find that CD and F-Score can be less reliable. When extracting surfaces via marching cubes, a level set threshold must be selected, which can introduce gaps relative to the ground truth due to the discrete grid resolution. This issue can disproportionately affect chunks of varying height, taller chunks may have fewer points per area given the same number of points are sampled for each chunk, leading to potentially higher CD and lower F-Score. As a result, these metrics are not fully comparable across settings like single-scene versus 4-scene training, due to differences in chunk distributions. In our case, larger differences in reported values are more indicative, while smaller variations may reflect noise due to the level set gap. Overall, IOU is more reliable and comparable across different scene configurations.

\mypara{Level of Detail.} In~\cref{fig:lod}, we visualize occupancy predictions at different resolutions during inference with marching cubes to generate different level of detail (LoD) for the triplane and vector set models. $50$ is the original chunk size used for training along the $x$ and $z$ axis, with the $y$ axis scaled according to the height prediction. We find that halving the chunk size offers comparable details while reducing memory usage for scenes. However, decreasing further to $13$ leads to noticeable loss of details in the trees and buildings.

\mypara{Additional Visualizations} We show additional results for our vecset diffusion model in~\cref{fig:more_3b61} and results for the RePaint baseline in~\cref{fig:more_3b61_repaint}.

\begin{figure*}
\centering
\includegraphics[width=0.95\linewidth]{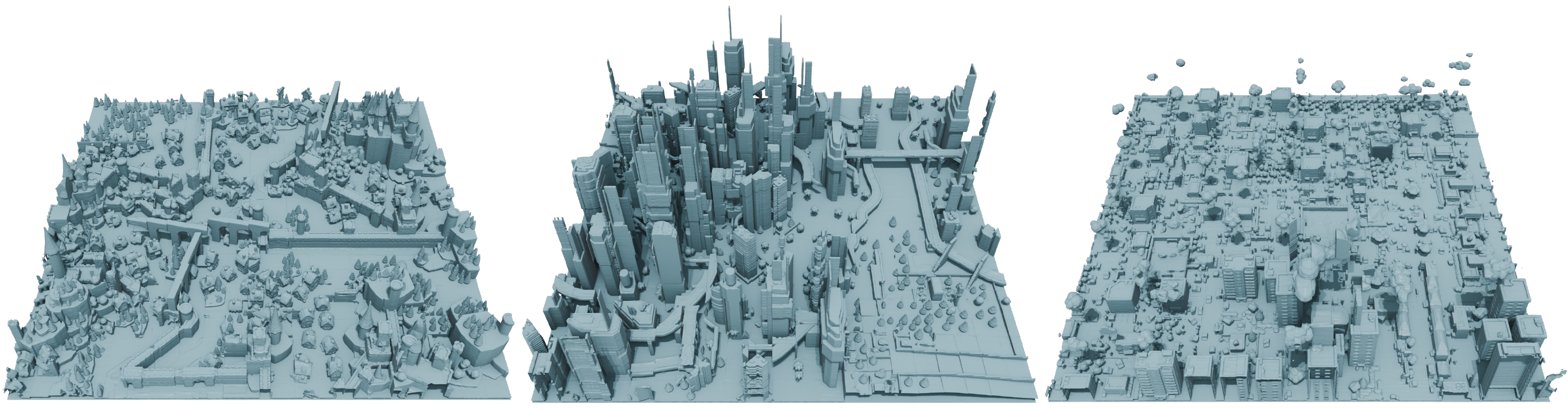}
\caption{Additional results for our vector set diffusion model trained on 4 scenes. The scenes shown here are of size $21\times21$.}
\label{fig:more_4_square}
\end{figure*}

\begin{figure*}
\centering
\includegraphics[width=0.95\linewidth]{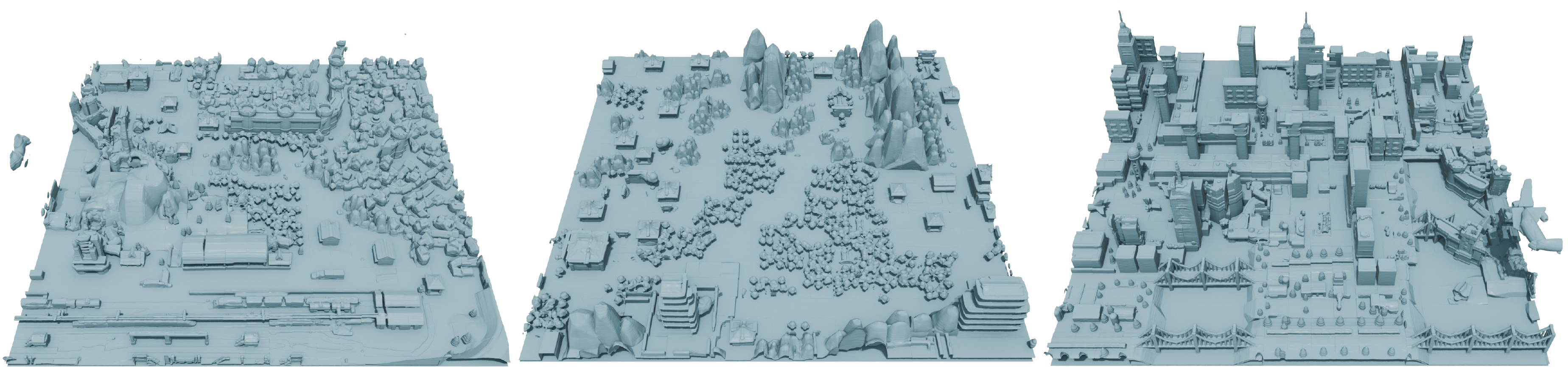}
\caption{Additional results for our vector set diffusion model trained on 13 scenes. The scenes shown here are of size $21\times21$.}
\label{fig:more_13_square}
\end{figure*}

\subsection{4-scene}

\begin{table}
\centering
\caption{Quantitative comparison of reconstruction across different VAE backbones on the 4 scene training. Here $\hat{h}$ indicates that the predicted height was used for the occupancy prediction and $\tilde{h}$ the ground truth height.}
\vspace{-1em}
\resizebox{\linewidth}{!}
{
\begin{tabular}{@{}llrrrrr@{}}
\toprule
Method & Output Res/S & IOU$\uparrow$ & CD $(\hat{h})\downarrow$ & F-Score $(\hat{h})\uparrow$ & CD $(\tilde{h})\downarrow$ & F-Score $(\tilde{h})\uparrow$\\
\midrule
triplane& $3\times64^2$/$6$ & 0.933 & 0.061 & 0.833 & 0.061 & 0.832 \\
vecset & - & $\mathbf{0.989}$ & $\mathbf{0.053}$ & $\mathbf{0.869}$ & $\mathbf{0.053}$ & $\mathbf{0.868}$ \\
\bottomrule
\end{tabular}
}
\label{tab:vae_recon_4_scene}
\end{table}

\begin{table}
\centering
\caption{Comparison of triplane and vecset diffusion models for generated quad-chunks on the 4 scene training. KPD scores are multiplied $10^3$.}
\vspace{-1em}
\resizebox{0.45\linewidth}{!}
{
\begin{tabular}{@{}lrr@{}}
\toprule
Method & FPD$\downarrow$ & KPD$\downarrow$ \\
\midrule
triplane & 0.868 & 2.142 \\
vecset & 0.581 & 1.104 \\
\bottomrule
\end{tabular}
}
\label{tab:kpd_fpd_4_scene}
\end{table}

For the 4-scene experiments in this section we follow the same GPU configurations in the main paper, and increase the number of sampled chunks across 4 scenes to 300K chunks. The training and validation follow the same 95\%-5\% split. For diffusion evaluation we sample 30K quad chunks from the scenes and well as the diffusion models for evaluation.

\mypara{Quantitative Results.} We show the VAE reconstruction and diffusion quality in~\cref{tab:vae_recon_4_scene} and~\cref{tab:kpd_fpd_4_scene}, respectively. We can see that for IOU the vector set model maintains its performance, while the triplane slightly drops. With the vector outperforming the triplane model across the board.

\mypara{Additional Visualizations.} We show additional results from our vecset diffusion model trained on 4 scenes in~\cref{fig:more_4_square}.

\subsection{13-scene}

To further demonstrate our model's potential for scaling up, we sample 280K chunks from 13 scenes in NuiScene43 (sample configurations available \href{https://github.com/3dlg-hcvc/NuiScene43-Dataset}{here}) and train using the same settings as before.

\mypara{Improving Reconstruction.} Following the previous vector set settings, the VAE model trained on 13 scenes see a drop in IOU to $0.968$, due to a larger variation of scenes compared to the single or 4-scene scenarios. One naive way we can increase the VAE's performance is to increase the number of feature vectors directly during training from $V=16$. However, this would increase the memory usage and slow diffusion training as shown in the triplane model. Instead, inspired by the deconvolution layers for triplanes. We introduce a pixel shuffle~\cite{shi2016real}-like layer for upsampling the number of vector sets followed by additional self attention layers. This increases the capacity of the model without increasing the latent size. Specifically, we add an additional projection layer that increases the number of channels and reshape from the channels dimension to the vector tokens. We upsample to $512$ vector sets and add 3 self-attention layers before the cross attention layer for querying coordinates. This improves the IOU to $0.983$.

\begin{figure}
\centering
\includegraphics[width=0.80\linewidth]{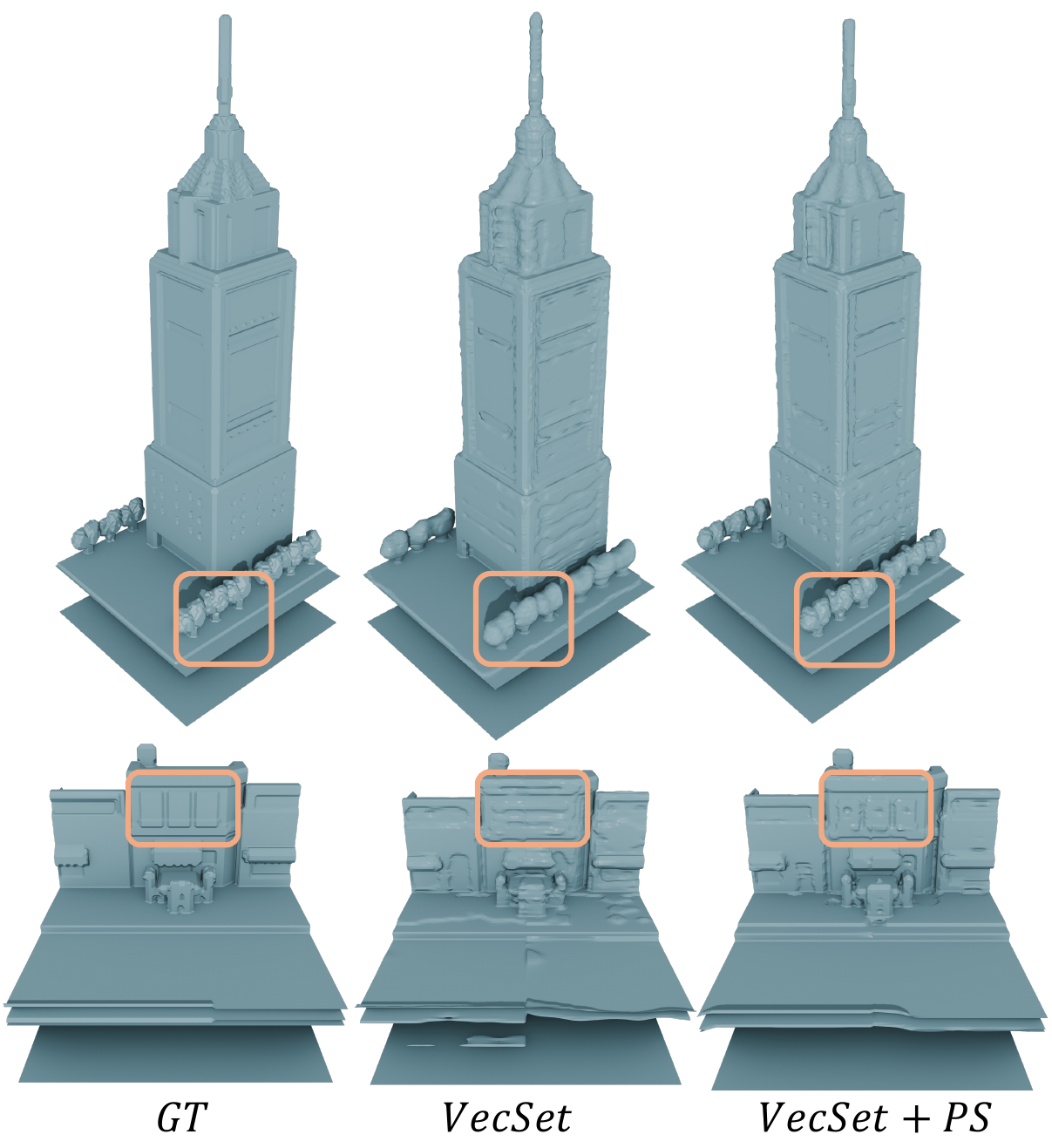}
\caption{We compare chunk reconstructions between the original vector set model and the upsampled version (VecSet + PS). The upsampling model recovers finer details that are blurry in the original (see orange boxes).}
\label{fig:13_vae_recon}
\end{figure}

In~\cref{fig:13_vae_recon}, we show a qualitative comparison between models. The original vector set model already reconstructs scenes quite well, despite the increase in the number of training scenes. The added improvements further enhance the reconstruction of finer details.

\mypara{Additional Visualizations.} We show square scenes generated by the diffusion model trained with the improved model with pixel shuffle in~\cref{fig:more_13_square}. And larger scenes with aerial and street view zoom-ins in~\cref{fig:13_scene_large}. It can be seen that this model can generate a larger variety of different scenes while maintaining good fidelity.

\end{document}